\PassOptionsToPackage{table}{xcolor}
\documentclass[]{lightspeed}

\usepackage[utf8]{inputenc}
\usepackage{latexsym}
\usepackage{fvextra}
\usepackage{tabularx}
\usepackage{array}
\usepackage{amsmath}
\usepackage{amssymb}
\usepackage{pifont}
\usepackage{wrapfig}

\microtypesetup{expansion=false}
\setcitestyle{square,comma,numbers,sort&compress}
\DeclareRobustCommand{\cmark}{{\large\ding{51}}}
\DeclareRobustCommand{\xmark}{{\large\ding{55}}}
\definecolor{lightgray}{gray}{0.92}

\title{Ex-Omni-2D: Expressive Omni-Modal Dialogue Models with Native Visual Presence}

\author[1,*]{Haoyu Zhang}
\author[2]{Zhipeng Li}
\author[1]{Xiaoying Tang}
\author[1,\dagger]{Tianshu Yu}
\author[3,\dagger]{Yiwen Guo}

\affiliation[1]{The Chinese University of Hong Kong, Shenzhen}
\affiliation[2]{LIGHTSPEED}
\affiliation[3]{Independent Researcher}

\contribution[*]{Work done during an internship at LIGHTSPEED}
\contribution[\dagger]{Corresponding authors}

\email{haoyuzhang3@link.cuhk.edu.cn}
\email{zhipengxli@tencent.com}
\email{xiaoyingtang@cuhk.edu.cn}
\email{yutianshu@cuhk.edu.cn}
\email{guoyiwen89@gmail.com}

\abstract{Omni-modal dialogue models can understand multimodal inputs and synthesize spoken replies, but a spoken answer still leaves the agent visually absent. We introduce \textbf{Ex-Omni-2D}, a framework that answers a multimodal query with coordinated text, personalized speech, and reference-conditioned video. The dialogue model first writes a structured \textit{Visual Thought Plan} (VTP) for scene, emotion, and motion, then generates the response text and multi-codebook speech units. These speech units are decoded into audio and aligned with video frames, giving the speech and avatar modules a common timing signal while allowing them to learn from different data sources. The video module is trained as a full-sequence Teacher conditioned on reference appearance, VTP semantics, and frame-aligned speech units. We further explore to distill it into a few-step block-causal \emph{Streaming Student}; its Prefix Streaming mechanism carries the previous clean latent into the next chunk and is analyzed as a partial mitigation for late-chunk subject drift. At $400\times720$/$720\times400$, the four-step four-GPU Student provides incremental output with lower startup latency than the full-sequence Teacher.}

\headercontent{%
  \begin{tabular}{c}
    \url{https://logo-cuhksz.github.io/Ex-Omni-2D}
  \end{tabular}%
}

\date{September 2026}

\begin{document}
\thispagestyle{firstheader}
\maketitle

\section{Introduction}

Omni-modal dialogue systems increasingly understand speech, text, and visual inputs and answer with natural spoken replies~\citep{DBLP:journals/corr/abs-2503-20215,DBLP:conf/iclr/FangGZMZ025,DBLP:conf/acl/FangZGZ025,DBLP:journals/corr/abs-2408-16725}. For many interactive settings, however, a spoken answer is only part of the response: a user also expects a visible agent whose appearance, affect, and motion fit the current exchange. We study this setting with \emph{dialogue-native video responses}: given a user query, reference image, and reference audio, the system generates response text, personalized speech, and a synchronized video of the referenced speaker.

Strong talking-avatar models already animate a portrait from speech and visual references~\citep{DBLP:journals/corr/abs-2512-19546,DBLP:journals/corr/abs-2508-18621,DBLP:journals/corr/abs-2506-18866,DBLP:journals/corr/abs-2508-08248,DBLP:journals/corr/abs-2505-20156}. They are less suited to dialogue by themselves because the visual behavior is usually controlled by a completed waveform and by an auxiliary prompt prepared outside the conversation in many systems. Joint audio-video generators provide tighter modality coupling~\citep{DBLP:conf/cvpr/RuanMYH0FYJG23,liu2025javisdit,Universe_1}, but they are typically prompt- or reference-driven content generators. The missing piece is a response interface that lets the dialogue model decide both what to say and how the speaking agent should look while saying it.

Ex-Omni-2D uses two lightweight interfaces for this purpose. A \textit{Visual Thought Plan} (VTP) records the scene, emotion, movement style, and motion details needed by the avatar generator. Multi-codebook speech units provide the timing and acoustic content used by both the speech decoder and the video generator. This design has a practical training advantage: speech and dialogue modules can learn from speech-centric data, while the avatar module can learn from filtered video clips annotated with VTPs and teacher-forced speech units. At inference time, the VTP and generated unit stream join these pathways into one response.

The video module is first trained as a full-sequence Teacher conditioned on the reference appearance, VTP semantics, and frame-aligned speech units, with bidirectional self-attention over the entire latent sequence. To reduce generation latency, we distill the Teacher into a block-causal Streaming Student, drawing inspiration from prior work \citep{Sun_2026_CVPR,gu2026anyflow}. The Student partitions the latent sequence into three-latent blocks, retaining bidirectional attention within each block while enforcing causal information flow across blocks through persistent historical KV caches. 

Extensive experiments shows that the Teacher gives the strongest visual quality, while the Streaming Student exposes the cost and benefit of incremental generation. Our contributions are summarized as follows:
\begin{itemize}
    \item We formulate omni-modal dialogue with dialogue-native, reference-conditioned video responses and introduce a structured VTP that exposes response-specific visual guidance.
    \item Building on pretrained multi-codebook speech units, we establish a shared acoustic-temporal interface that supports personalized speech generation and frame-aligned avatar-video conditioning. This interface enables the dialogue-to-speech and speech-to-video pathways to learn from heterogeneous, pathway-specific datasets without requiring fully paired dialogue–speech–video response data.
    \item We explore a deployment-oriented Streaming Student distilled from the full-sequence Teacher, report its quality--latency trade-off under few-step incremental inference, and analyze prefix streaming as a partial mitigation for late-chunk drift.
\end{itemize}

\section{Related Work}
\label{sec:related_work}

\paragraph{Omni-modal Dialogue.}
Recent omni-modal large language models extend text-based interaction toward unified understanding and generation over speech, vision, and language. Representative systems explore speech-centered instruction following, real-time spoken response generation, cross-modal alignment, and unified multimodal generation within LLM-based architectures~\citep{DBLP:journals/corr/abs-2408-16725,DBLP:journals/corr/abs-2410-11190,luo2025openomni,DBLP:journals/corr/abs-2503-20215,DBLP:journals/corr/abs-2506-09344,fu2026vita,DBLP:journals/corr/abs-2511-00279}. In parallel, speech language models further reduce the dependence on cascaded ASR--LLM--TTS pipelines by directly modeling spoken dialogue and speech generation~\citep{DBLP:conf/acl/AoWZ0RW0KLZWQ0W22,DBLP:journals/corr/abs-2401-13527,DBLP:journals/corr/abs-2410-00037,DBLP:journals/corr/abs-2412-02612}. However, most existing omni-modal dialogue systems still lack visual presence: they can understand speech and synthesize spoken responses, but do not directly generate a speaking visual agent. 

\begin{figure*}[ht]
\begin{center}
    \includegraphics[width=0.70\linewidth]{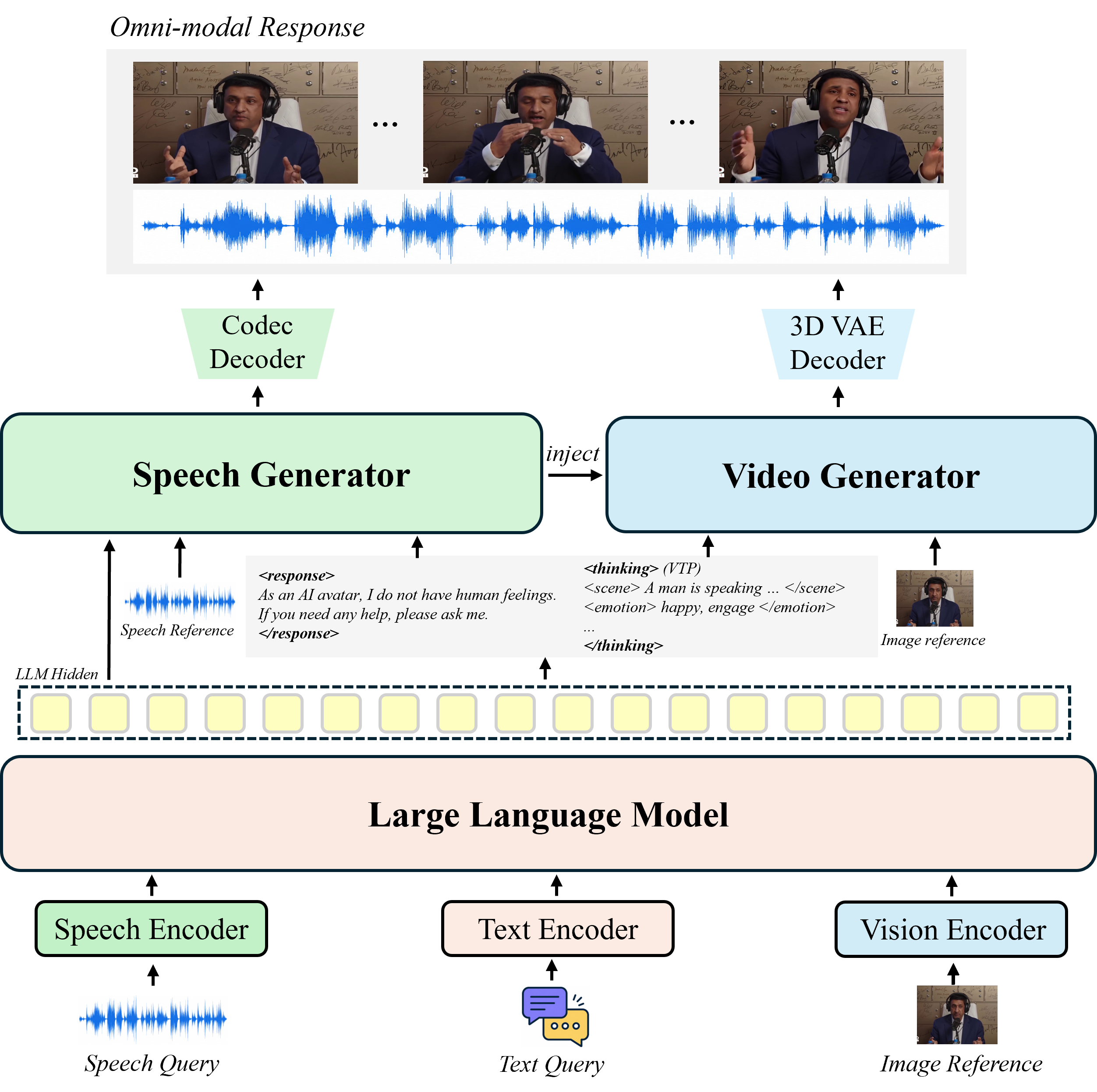}
\end{center}
    \caption{Overview of Ex-Omni-2D. Speech, text, and vision encoders provide multimodal context to the LLM, which produces a non-user-facing VTP and the user-facing response. The Speech Generator uses reference audio for voice conditioning and generates speech in a native multi-codebook space. The Video Generator encodes the reference image and VTP, and reuses the generated speech representation to synthesize the synchronized video response.}
\label{fig:pipeline}
\end{figure*}

\paragraph{Unified Audio-Video Generation.}
Aligned audio-video generation has been studied through both cascaded and joint paradigms. Audio-driven video synthesis first generates or receives a speech waveform and then animates a reference identity, which has led to strong progress in talking-head and human animation systems~\citep{10.1145/3394171.3413532,Zhang_2023_CVPR,xu2024hallohierarchicalaudiodrivenvisual,DBLP:journals/corr/abs-2506-18866,DBLP:journals/corr/abs-2508-18621}. The reverse direction, video-to-audio dubbing, generates sound or speech for a given silent video~\citep{NEURIPS2023_98c50f47,Cheng_2025_CVPR,DBLP:journals/corr/abs-2505-20156,Sung-Bin_2025_ICCV}. These conditional systems are effective visual or acoustic renderers, but their inputs usually do not include the conversational state from which response-specific visual behavior should be derived. Joint audio-video generation instead synthesizes both modalities within a unified or interacting generative process. Early diffusion-based work such as MM-Diffusion \citep{DBLP:conf/cvpr/RuanMYH0FYJG23} initiated this direction, followed by models that improve scalability and synchronization such as SyncFlow \citep{liu2024syncflowtemporallyalignedjoint} and UniForm~\citep{zhao2025uniformunifiedmultitaskdiffusion}. More recent systems, including JavisDiT \citep{liu2025javisdit}, Ovi \citep{low2025ovitwinbackbonecrossmodal}, UniVerse-1 \citep{Universe_1}, and UniAVGen \citep{zhang2026uniavgenunifiedaudiovideo}, further explore large-scale or human-centric joint audio-video generation~\citep{liu2025javisdit,low2025ovitwinbackbonecrossmodal,Universe_1,zhang2026uniavgenunifiedaudiovideo}. UniVerse-1 stitches pre-trained audio and video experts to produce coordinated audio-video content, while UniAVGen introduces a dual-branch DiT with asymmetric cross-modal interactions and face-aware modulation for human-centric generation. These methods are the closest comparison group to our framework because they generate paired audio and video. Ex-Omni-2D connects this generation problem to dialogue: the model first plans visual behavior from the current multimodal query, then passes both semantic guidance and acoustic timing to the avatar generator.

\section{Method}
\label{sec:method}
Ex-Omni-2D receives a multimodal user query $x=(x_t,x_s)$ containing text, speech, or both, together with a reference image $I^{\mathrm{ref}}$ and reference audio $a^{\mathrm{ref}}$. It generates response text $\mathbf{y}$, personalized speech, and a reference-conditioned video response $V$. Figure~\ref{fig:pipeline} shows the two interfaces that connect these outputs. The VTP $\mathbf{p}$ carries high-level visual intent, and the multi-codebook speech units $\mathbf{U}$ carry the acoustic content and timing shared by speech and video. The dialogue model emits the VTP and response text; response-side hidden states drive the Speech Generator; the resulting units are decoded into waveform audio and mapped to frame-aligned video conditions. The Video Generator has two realizations: a full-sequence Teacher with bidirectional temporal context and a Prefix-Streaming Student that renders chunks incrementally while carrying motion state forward.

\subsection{Multimodal Character Grounding}
\label{sec:method_input}

Let $x_t$ and $x_s$ denote the text and speech components of the current query, where either component may be empty but at least one is present. The text is embedded by the language model, while speech is encoded and projected into the same hidden space:
\begin{equation}
    \mathbf{E}_t = \mathrm{Emb}_{\mathrm{text}}(x_t), \quad
    \mathbf{E}_s = P_{\mathrm{sp}}\!\left(E_{\mathrm{sp}}(x_s)\right).
\end{equation}
The two sequences are combined according to their order in the query to form the multimodal query representation $\mathbf{E}_x$.

The character conditions follow distinct pathways. For dialogue understanding and VTP generation, the reference image is processed by the vision tower of Qwen3-VL-2B \citep{bai2025qwen3} and projected into the dialogue-model space:
\begin{equation}
    \mathbf{z}^{\mathrm{llm}}_I
    = P_{\mathrm{vis}}\!\left(E_{\mathrm{vis}}(I^{\mathrm{ref}})\right).
\end{equation}
Together, $\mathbf{E}_x$ and $\mathbf{z}^{\mathrm{llm}}_I$ form the dialogue-model input. The same image is independently encoded inside the Video Generator by the Wan 3D VAE \citep{wan2025} to obtain an appearance-reference latent:
\begin{equation}
    \mathbf{R}^{\mathrm{ref}}
    = E_{\mathrm{vae}}\!\left(I^{\mathrm{ref}}\right).
\end{equation}
Reference audio is handled inside the Speech Generator, where a speaker encoder extracts the voice embedding $\mathbf{s}^{\mathrm{ref}}$. 

\subsection{Visual Response Planning}
\label{sec:method_llm}

The dialogue backbone is instantiated with Qwen3-8B~\citep{yang2025qwen3}. It follows a structured assistant protocol that separates the internal visual plan from the user-facing response:
\begin{equation}
\begin{split}
    \mathbf{o} = [&\texttt{<thinking>}, \mathbf{p},
    \texttt{</thinking>},
    \texttt{<response>}, \mathbf{y},
    \texttt{</response>}].
\end{split}
\end{equation}
The \texttt{<thinking>} block is restricted to the structured VTP rather than an unconstrained chain-of-thought trace. Atomic boundary tokens and constrained decoding ensure that the VTP is completed before response generation and that the planning and user-facing spans can be routed independently. The VTP contains five fields:
\begin{equation}
    \mathbf{p} =
    (p_{\mathrm{first}},p_{\mathrm{scene}},p_{\mathrm{emotion}},
    p_{\mathrm{style}},p_{\mathrm{motion}}),
\end{equation}
which describe the first-frame scene, overall scene, emotion, movement style, and detailed motion. The plan is internal and is not displayed as the conversational answer. The Video Generator encodes it into a semantic condition
\begin{equation}
    \mathbf{L}^{\mathrm{vtp}}
    = E_{\mathrm{text}}\!\left(\mathbf{p}\right).
\end{equation}
Because the plan is text, it can be supervised with ordinary autoregressive training and inspected during evaluation. We retain the final-layer hidden states only over the response span, denoted $\mathbf{H}^{\ell}_{\mathbf{y}}$, for speech generation. The VTP provides high-level semantic guidance for how the response should look, while $\mathbf{H}^{\ell}_{\mathbf{y}}$ and $\mathbf{y}$ specify what should be spoken. The final video realization is jointly conditioned on the VTP and frame-aligned speech units.

\subsection{Speech Generator and Shared Acoustic Interface}
\label{sec:method_speech}

The Speech Generator is initialized from Qwen3-TTS-0.6B~\citep{hu2026qwen3ttstechnicalreport}. Conditioned on the response states, response tokens, and reference voice, it predicts a sequence of multi-codebook acoustic units:
\begin{equation}
    \mathbf{U}
    = G_{\mathrm{sp}}(\mathbf{H}^{\ell}_{\mathbf{y}},
    \mathbf{y},\mathbf{s}^{\mathrm{ref}}),
    \quad \mathbf{U}\in\mathbb{N}^{N\times C},
\end{equation}
where $C=16$ is the number of Qwen3-TTS acoustic codebooks. The first codebook is generated autoregressively at each acoustic frame, and the residual codebooks refine its acoustic content. Text-token embeddings and $\mathbf{H}^{\ell}_{\mathbf{y}}$ are fused through gated cross-attention before acoustic decoding.

We treat $\mathbf{U}$ as the shared representation of the generated speech. A codec decoder renders it as waveform audio, while the avatar pathway maps the same units into video-conditioning features. For acoustic frame $n$, the codebook embeddings are aggregated and projected by a lightweight adapter:
\begin{equation}
    \widetilde{\mathbf{A}}_n
    = f_{\mathrm{aud}}\!\left(
    \sum_{c=1}^{C} E_c(u_{n,c})\right).
\end{equation}
The units are produced at $12.5$ Hz and the video at $25$ FPS, so each acoustic feature is repeated for two video frames:
\begin{equation}
    \mathbf{A}_{2n-1}=\mathbf{A}_{2n}
    =\widetilde{\mathbf{A}}_n,
    \quad n=1,\ldots,N,
\end{equation}
where $\mathbf{A}\in\mathbb{R}^{2N\times d_a}$ is aligned with the video timeline. This interface uses all 16 codebooks and gives the video generator an explicit frame-level timing signal without re-encoding a completed waveform. It also makes the training data easier to use: speech modeling can use speech and spoken-dialogue data, while the video pathway can use avatar clips with teacher-forced speech units. At inference time, the video pathway receives the dialogue-derived VTP together with the same generated acoustic representation used to render the spoken response.

\subsection{Video Generator}
\label{sec:method_video}
\paragraph{Full-Sequence Teacher.}
Ex-Omni-2D first trains a full-sequence Video Generator based on Wan2.1-T2V-1.3B~\citep{wan2025} and initialized with the corresponding OmniAvatar-1.3B LoRA weights~\citep{DBLP:journals/corr/abs-2506-18866}. It contains a VTP text encoder, a 3D VAE, and a latent denoising backbone, and uses bidirectional temporal context over the target clip. This full-sequence realization constitutes the primary Video Generator and later serves as the Teacher for the distillation stage. The Teacher receives three complementary conditions: the reference latent $\mathbf{R}^{\mathrm{ref}}$ for appearance, the frame-aligned acoustic condition $\mathbf{A}$ for speech-driven motion, and the VTP condition $\mathbf{L}^{\mathrm{vtp}}$ for scene, affect, and movement semantics:
\begin{equation}
    V = G_{\mathrm{vid}}(\mathbf{R}^{\mathrm{ref}},
    \mathbf{A},\mathbf{L}^{\mathrm{vtp}}).
\end{equation}

For a ground-truth video $V^{\ast}$, the 3D VAE produces a clean latent $\mathbf{x}_0$. Following the flow-matching formulation of~\citet{gu2026anyflow}, We sample Gaussian noise $\boldsymbol{\epsilon}$ and a noise level $\sigma$ and construct the flow-matching input
\begin{equation}
    \mathbf{x}_{\sigma}
    =(1-\sigma)\mathbf{x}_0+\sigma\boldsymbol{\epsilon}.
\end{equation}
The denoising network is conditioned on $\mathbf{x}_{\sigma}$, $\sigma$, $\mathbf{R}^{\mathrm{ref}}$, $\mathbf{A}$, and $\mathbf{L}^{\mathrm{vtp}}$. After alignment to the 25-FPS timeline, the acoustic features are packed to the video-latent temporal resolution. Following OmniAvatar~\citep{DBLP:journals/corr/abs-2506-18866}, the projected speech features are injected into DiT blocks 2--15. At an injected layer $\ell$, the aligned feature is tiled over the latent spatial grid, patchified, and added to the video tokens:
\begin{equation}
    \mathbf{X}^{(\ell)}
    \leftarrow \mathbf{X}^{(\ell)}+\mathbf{G}^{(\ell)}.
\end{equation}

\paragraph{Streaming Student.}
\label{sec:method_streaming}

We also explore to distill the full-sequence Teacher into a block-causal Streaming Student that renders speech-aligned video chunks incrementally. The Student uses the generated acoustic units as the timing signal and carries the previous chunk's final clean latent as a one-latent prefix, giving each new chunk a local motion boundary condition while keeping future frames unavailable. The full distillation objective, Prefix-Streaming state update, and rate-aligned chunk construction are described in Appendix~\ref{sec:streaming_student_details}.

\subsection{Training Objectives and Stages}
\label{sec:method_objectives}

The core Ex-Omni-2D framework uses three pathway-specific supervision signals: an autoregressive language loss over the VTP and response protocol, a multi-codebook speech-generation loss conditioned on the response span, and a flow-matching video loss conditioned on the reference image, teacher-forced speech units, and VTP. For efficient incremental deployment, the frozen Teacher then supervises the Streaming Student through the flow-map and on-policy distribution-matching procedure described in Section~\ref{sec:method_streaming} and Appendix~\ref{sec:training_details}.

Training uses three core stages followed by one deployment-oriented distillation stage. \textbf{Stage 1: Speech Interface Alignment} uses approximately 800K ASR and 1M TTS examples to train the Speech Projector and Speech Generator while keeping the LLM frozen. \textbf{Stage 2: Omni-modal Response Adaptation} updates the LLM, Speech Projector, and Speech Generator with InstructS2S-200K~\citep{DBLP:conf/iclr/FangGZMZ025} and OmniCharacter~\citep{zhang-etal-2025-omnicharacter}, introducing the structured VTP--response protocol and adapting the shared speech interface to interactive responses. Because these dialogue datasets do not contain target response videos, Qwen3-VL-235B-A22B-Instruct constructs the target VTP from the reference image, dialogue context, and target response using the five-field template in Appendix~\ref{sec:prompt_templates}. \textbf{Stage 3: Avatar Video Realization} operates on about 140K filtered SpeakerVid clips. Each clip is converted into an avatar-response record containing a reference frame, a five-field VTP annotation, aligned multi-codebook speech units, and the target video. These records train the full-sequence Video Generator to realize the semantic and acoustic-temporal interfaces produced by the response pathway. The resulting model is frozen as the Teacher. \textbf{Stage 4: Streaming Student Distillation} then transfers the Teacher into the deployment-oriented Student through Phase~I flow-map learning and Phase~II on-policy distribution matching. At inference time, VTP and speech units reconnect the pathways without requiring large-scale dialogues paired with response videos. Data construction, optimization, and inference configurations are provided in Appendix~\ref{sec:training_details}.

\begin{table*}[htbp]
  \centering
  \caption{Capability-oriented audio-video comparison across three generation paradigms under the protocol in Section~\ref{sec:eval_metrics}. A dash denotes a metric not applicable to the supported interface. Identical capability indicators, parameter counts, and audio metrics are merged within the Ex-Omni-2D block. }
  \label{tab:main_results}
  \scriptsize
  \setlength{\tabcolsep}{2.75pt}
  \renewcommand{\arraystretch}{1.15}
  \begin{tabularx}{\textwidth}{lcccccccccccc}
    \toprule
    \multirow{2}{*}{\textbf{Method}} & \multirow{2}{*}{\textbf{Dialogue}} & \multirow{2}{*}{\textbf{Ref. Voice}} & \multicolumn{3}{c}{\textbf{Parameters}} & \multicolumn{3}{c}{\textbf{Audio Quality}} & \multicolumn{3}{c}{\textbf{Video Quality}} & \multicolumn{1}{c}{\textbf{A-V Sync}} \\
    \cmidrule(lr){4-6} \cmidrule(lr){7-9} \cmidrule(lr){10-12} \cmidrule(lr){13-13}
    & & & \textbf{LLM} & \textbf{Speech Gen.} & \textbf{Video Gen.} & \textbf{PQ} $\uparrow$ & \textbf{CU} $\uparrow$ & \textbf{SIM} $\uparrow$ & \textbf{SC} $\uparrow$ & \textbf{IQ} $\uparrow$ & \textbf{DD} & \textbf{Sync-C} $\uparrow$ \\
    \midrule
    \addlinespace[-0.05pt]
    \rowcolor{lightgray} \multicolumn{13}{c}{Video Generation} \\
    {echomimic} & \xmark & \xmark & - & - & 1.3B & - & - & - & \textbf{97.87} & 54.37 & \textbf{74.00} & 4.82 \\
    {StableAvatar-1.3B} & \xmark & \xmark & - & - & 1.3B & - & - & - & 97.40 & 67.51 & 42.00 & 3.38 \\
    {OmniAvatar-1.3B} & \xmark & \xmark & - & - & 1.3B & - & - & - & \underline{97.74} & 66.65 & 15.50 & \textbf{5.64} \\
    {FantasyTalking} & \xmark & \xmark & - & - & 1.3B & - & - & - & 96.78 & 64.21 & 6.50 & 3.43 \\
    \midrule
    \addlinespace[-0.05pt]
    \rowcolor{lightgray} \multicolumn{13}{c}{Video-Audio Generation} \\
    {Universe-1} & \xmark & \xmark & - & 3.5B & 1.3B & 4.16 & 3.98 & - & 97.35 & \underline{67.72} & 34.00 & 1.75 \\
    {UniAVGen} & \xmark & \cmark & - & 1.4B & 5B & - & - & 0.291 & 96.07 & \textbf{68.02} & 55.00 & 4.15 \\
    \midrule
    \addlinespace[-0.05pt]
    \rowcolor{lightgray} \multicolumn{13}{c}{Dialogue-Based Omni-Modal Response Generation} \\
    Ex-Omni-2D (Teacher) & \multirow{4}{*}{\cmark} & \multirow{4}{*}{\cmark} & \multirow{4}{*}{8B} & \multirow{4}{*}{0.6B} & \multirow{4}{*}{1.3B} & \multirow{4}{*}{\textbf{7.53}} & \multirow{4}{*}{\textbf{7.06}} & \multirow{4}{*}{\textbf{0.417}} & 94.62 & 67.31 & \underline{72.00} & \underline{4.95} \\
    Ex-Omni-2D (Streaming Stu, 2-step) & & & & & & & & & 93.33 & 52.70 & 9.50 & 3.51 \\
    Ex-Omni-2D (Streaming Stu, 4-step) & & & & & & & & & 93.65 & 57.40 & 32.00 & 3.90 \\
    Ex-Omni-2D (Streaming Stu, 8-step) & & & & & & & & & 93.91 & 61.15 & 48.00 & 4.00 \\
    \bottomrule
  \end{tabularx}

\end{table*}

\section{Experiments}
\label{sec:experiments}

\subsection{Evaluation}
\label{sec:eval_metrics}
\paragraph{Benchmarks.}
We evaluate audio-video generation on the 200-query CommonEval speech-QA split from VoiceBench~\citep{chen2024voicebenchbenchmarkingllmbasedvoice}, pairing each query with fixed reference image and speech conditions sampled from the held-out SpeakerVid test split~\citep{zhang2025speakervid}. Spoken question answering uses AlpacaEval, CommonEval, and BBH from VoiceBench; multi-turn dialogue quality uses the 400-dialogue OmniCharacter benchmark~\citep{zhang-etal-2025-omnicharacter}. For cascaded audio-video baselines, Qwen2.5-Omni-7B~\citep{DBLP:journals/corr/abs-2503-20215} provides the shared upstream response, while each method is evaluated through the modalities supported by its public interface. Full baseline lists and modality-specific comparison details are provided in Appendix~\ref{sec:eval_metric_details}.

\paragraph{Metrics.}
We report audio quality (PQ, CU, SIM), video quality (SC, IQ, DD), audio--video synchronization (Sync-C), VoiceBench speech-QA scores, and OmniCharacter dialogue scores. Detailed metric definitions and scoring protocols are provided in Appendix~\ref{sec:eval_metric_details}.

\subsection{Audio-Video Comparison Results}
\label{sec:exp_results}
Table~\ref{tab:main_results} compares video generation, joint video-audio generation, and dialogue-based omni-modal response generation under the protocol above. Among the downstream baselines, echomimic obtains the highest subject consistency (SC 97.87) and largest motion incidence by DD (74.00), UniAVGen obtains the highest imaging quality (IQ 68.02), and OmniAvatar-1.3B obtains the strongest synchronization (Sync-C 5.64). The Ex-Omni-2D Teacher does not dominate these standalone rendering metrics; instead, it obtains SIM 0.417, DD 72.00, and Sync-C 4.95 while additionally generating the dialogue response, personalized speech, and query-derived visual plan. Increasing the Streaming Student's denoising budget from two to eight steps consistently increases SC, IQ, motion incidence, and Sync-C, exposing a controllable quality--efficiency trade-off. Appendix~\ref{sec:student_denoising_steps} analyzes this trade-off in detail, while Appendix Table~\ref{tab:upstream_cascade_ablation} studies the effect of replacing the cascaded baselines' upstream controller with Ex-Omni-2D.

\begin{table*}[htbp]
\centering
\caption{Full multi-turn dialogue evaluation on OmniCharacter. We follow OmniCharacter's official evaluation protocol. KE: Knowledge-Exposure, KA: Knowledge-Accuracy, KH: Knowledge-Hallucination, PB: Persona-Behavior, PU: Persona Utterance, Flu.: Fluency, Coh.: Coherency, Cons.: Consistency, HL: Human-Likeness, CS: Communication Skill, ED: Expression Diversity, and Emp.: Empathy.}
\label{tab:omnicharacter_comparison}
\setlength{\tabcolsep}{5.3pt}
\renewcommand{\arraystretch}{1.05}
\scriptsize
\begin{tabularx}{\textwidth}{lccccccccccccc}
\toprule
\multirow{2}{*}{\textbf{Model}}
& \multicolumn{5}{c}{\textbf{Character Consistency}}
& \multicolumn{3}{c}{\textbf{Conversational Ability}}
& \multicolumn{4}{c}{\textbf{Role-playing Attractiveness}}
& \multirow{2}{*}{\textbf{Avg.} $\uparrow$} \\
\cmidrule(lr){2-6}\cmidrule(lr){7-9}\cmidrule(lr){10-13}
& \textbf{KE} $\uparrow$ & \textbf{KA} $\uparrow$ & \textbf{KH} $\uparrow$ & \textbf{PB} $\uparrow$ & \textbf{PU} $\uparrow$
& \textbf{Flu.} $\uparrow$ & \textbf{Coh.} $\uparrow$ & \textbf{Cons.} $\uparrow$
& \textbf{HL} $\uparrow$ & \textbf{CS} $\uparrow$ & \textbf{ED} $\uparrow$ & \textbf{Emp.} $\uparrow$ & \\
\midrule
ChatGLM3-6B
& 2.016 & 2.792 & 2.704 & 2.455 & 2.812
& 3.269 & 3.647 & 3.283
& 3.064 & 2.932 & 1.969 & 2.993 & 2.828 \\
Baichuan2-7B
& 1.813 & 2.849 & 2.929 & 2.830 & \underline{3.081}
& 3.551 & 3.894 & \underline{3.827}
& 3.670 & 2.728 & 2.115 & 2.984 & 3.023 \\
InternLM-7B
& 1.782 & 2.800 & 2.781 & 2.719 & 3.016
& 3.527 & 3.823 & 3.744
& 3.546 & 2.622 & 2.070 & 2.897 & 2.944 \\
CharacterGLM-6B
& 1.640 & 2.819 & 2.738 & 2.301 & 2.969
& 3.414 & 3.717 & 3.737
& \underline{3.738} & 2.265 & 1.966 & 2.937 & 2.853 \\
Llama-3.1-8B
& 2.197 & 2.701 & 2.615 & 3.130 & 2.704
& 3.059 & 3.477 & 3.071
& 2.922 & 2.934 & 2.634 & 2.759 & 2.850 \\
Qwen-7B
& 1.956 & 2.728 & 2.633 & 2.605 & 2.780
& 3.187 & 3.564 & 3.229
& 3.036 & 2.791 & 2.052 & 2.838 & 2.783 \\
Qwen2-7B-Instruct
& 1.966 & 2.537 & 2.412 & 2.313 & 2.436
& 2.864 & 3.171 & 2.743
& 2.655 & 2.612 & 1.867 & 2.654 & 2.519 \\
Qwen2.5-7B-Instruct
& 2.172 & 3.012 & 2.887 & 3.526 & 2.879
& 3.321 & 3.657 & 3.326
& 3.278 & 3.127 & 2.728 & 3.217 & 3.094 \\
Qwen3-8B
& \textbf{2.722} & 3.199 & \underline{2.950} & \textbf{3.565} & 3.044
& 3.497 & 3.795 & 3.318
& 3.148 & \textbf{3.566} & \textbf{3.030} & 3.338 & \underline{3.264} \\
\midrule
OmniCharacter
& 2.230 & 3.040 & 2.918 & \underline{3.531} & 2.988
& 3.369 & 3.768 & 3.410
& 3.374 & 3.261 & \underline{3.002} & 3.187 & 3.173 \\
Qwen2.5-Omni-7B
& \underline{2.462} & \underline{3.266} & 2.943 & 3.148 & \underline{3.070}
& 3.692 & \underline{3.951} & 3.617
& 3.486 & \underline{3.481} & 2.502 & \textbf{3.474} & 3.258 \\
\midrule
\textbf{Ex-Omni-2D}
& 2.076 & \textbf{3.410} & \textbf{2.994} & 2.911 & \textbf{3.354}
& \textbf{3.812} & \textbf{4.100} & \textbf{3.902}
& \textbf{3.893} & 3.362 & 2.239 & \underline{3.341} & \textbf{3.283} \\
\bottomrule
\end{tabularx}
\end{table*}

\subsection{Multi-turn Dialogue Quality}
\label{sec:dialogue_results}

We use OmniCharacter~\citep{zhang-etal-2025-omnicharacter} primarily to evaluate multi-turn dialogue quality: whether the model remains fluent, coherent, and consistent when responding with access to the preceding conversation. Following the benchmark's complete official protocol, Table~\ref{tab:omnicharacter_comparison} also includes its character-consistency and role-playing-attractiveness dimensions. Ex-Omni-2D obtains the highest reported Fluency, Coherency, and Consistency values among the compared methods, averaging 3.938 across the three metrics compared with 3.537 for its Qwen3-8B backbone. Its twelve-dimension average is 3.283, close to Qwen3-8B at 3.264; the metric-wise results include both increases and decreases relative to Qwen3-8B, showing that visual-response training changes the capability profile rather than uniformly improving every dimension. Appendix Table~\ref{tab:omnicharacter_context_ablation} directly evaluates the effect of providing dialogue history.

\begin{table}[htbp]
\centering

\begin{minipage}[htbp][3.8cm][t]{0.49\textwidth}
\centering

\caption{Speech QA comparison. $^*$ indicates results reproduced
using open-source code.}
\label{tab:speech_qa}

\vfill

\scriptsize
\setlength{\tabcolsep}{8.5pt}
\renewcommand{\arraystretch}{1.08}
\begin{tabularx}{\linewidth}{lccc}
\toprule
\textbf{Model}
& \textbf{AlpacaEval} $\uparrow$
& \textbf{CommonEval} $\uparrow$
& \textbf{BBH} $\uparrow$ \\
\midrule

Qwen2.5-Omni
& \textbf{4.49}
& \textbf{3.93}
& \textbf{60.80$^*$} \\

Moshi
& 2.01
& 1.60
& 47.40 \\

VITA-1.5
& 4.21
& 3.66
& 55.30 \\

Mini-Omni2
& 2.32
& 2.18
& 46.40 \\

SLAM-Omni
& 1.90
& 1.79
& 48.80 \\
\midrule

Ex-Omni-2D
& \underline{4.28}
& \underline{3.71}
& \underline{58.70} \\
\bottomrule
\end{tabularx}

\end{minipage}
\hfill
\begin{minipage}[htbp][3.8cm][t]{0.49\textwidth}
\centering
\caption{Quality--efficiency Comparison between Teacher and
Streaming Student. Efficiency is measured at
$400\times720$/$720\times400$.}
\label{tab:streaming_latency}

\scriptsize
\setlength{\tabcolsep}{3.5pt}
\renewcommand{\arraystretch}{1.28}
\begin{tabularx}{\linewidth}{
    @{}>{\raggedright\arraybackslash}l c c c c c c c@{}
}
\toprule
\textbf{Model}
& \textbf{Steps}
& \textbf{SC} $\uparrow$
& \textbf{IQ} $\uparrow$
& \textbf{DD} 
& \textbf{Sync-C} $\uparrow$
& \textbf{FPS} $\uparrow$
& \textbf{E2E RTF} $\downarrow$ \\
\midrule

Teacher
& 50
& \textbf{94.62}
& \textbf{67.31}
& \textbf{72.00}
& \textbf{4.95}
& 1.409
& 26.917 \\
\midrule

Student
& 2
& 93.33
& 52.70
& 9.50
& 3.51
& \textbf{39.546}
& \textbf{1.201} \\

Student
& 4
& 93.65
& 57.40
& 32.00
& 3.90
& 26.512
& 1.293 \\

Student
& 8
& 93.91
& 61.15
& 48.00
& 4.00
& 15.622
& 1.932 \\
\bottomrule
\end{tabularx}
\end{minipage}

\end{table}

\subsection{Speech QA Comparison Results}
\label{sec:speech_qa_results}
As shown in Table \ref{tab:speech_qa}, on VoiceBench~\citep{chen2024voicebenchbenchmarkingllmbasedvoice}, Ex-Omni-2D obtains 4.28 on AlpacaEval, 3.71 on CommonEval, and 58.70 on BBH. Within the listed systems, these are the second-highest reported scores after Qwen2.5-Omni. In addition, Appendix~\ref{sec:additional_dialogue_results} discuss the impact of VTP on the performance of question-answering.

\subsection{Quality--Efficiency Trade-off}
\label{sec:streaming_latency}
Table~\ref{tab:streaming_latency} compares the Teacher and Streaming Student under the four-GPU deployment. The Teacher achieves the highest values across all reported metrics, but runs at only 1.4 FPS with an end-to-end RTF of 26.9. For the Student, increasing the denoising steps consistently improves SC, IQ, DD, and Sync-C, at the cost of lower rendering speed. The two-step Student is the fastest at 39.5 FPS, but shows a substantial quality drop. The eight-step
Student recovers noticeably better quality and motion, while still reaching 15.6 FPS. The four-step Student provides a favorable intermediate point, running the video renderer at 26.5 FPS with an end-to-end RTF of 1.29. Although the two- and four-step renderers exceed the 25-FPS playback rate, the complete pipeline remains slightly slower than real time. This gap arises because end-to-end latency also includes autoregressive response generation, speech-unit generation, and synchronization between the speech and video stages. Full results across GPU settings and resolutions are reported in Appendix Table~\ref{tab:streaming_latency_full}, with additional analyses of the Prefix-Streaming Student in Appendix~\ref{sec:prefix_comparison}.

\subsection{Effect of VTP and Reference Speech}
\label{sec:vtp_ablation}

We ablate response-specific VTP and personalized reference speech on the same 200 CommonEval samples used in Table~\ref{tab:main_results}. We replace the generated VTP with a fixed neutral plan and the matched reference speech with a fixed public utterance. As shown in Table~\ref{tab:vtp_reference_speech_full}, removing the response-specific VTP consistently lowers SC and Sync-C, demonstrating the benefit of planning visual responses according to the dialogue context. Although DD increases in some settings, DD measures motion magnitude rather than motion quality or semantic appropriateness. Using the public reference utterance improves PQ and CU but reduces SIM from 0.417 to 0.015, indicating that these gains come from the
selected voice at the cost of speaker personalization.

\begin{table*}[htbp]
  \centering
  \caption{Complete ablation of response-specific VTP and personalized reference speech. SIM is evaluated against each sample's original matched reference speech.}
  \label{tab:vtp_reference_speech_full}
  \scriptsize
  \setlength{\tabcolsep}{7.0pt}
  \renewcommand{\arraystretch}{1.12}
  \begin{tabularx}{\textwidth}{lccccccccc}
    \toprule
    \textbf{Setting} & \textbf{Specific VTP} & \textbf{Personalized Speech} & \textbf{PQ} $\uparrow$ & \textbf{CU} $\uparrow$ & \textbf{SIM} $\uparrow$ & \textbf{SC} $\uparrow$ & \textbf{IQ} $\uparrow$ & \textbf{DD} & \textbf{Sync-C} $\uparrow$ \\
    \midrule
    Full & \cmark & \cmark & 7.53 & 7.06 & \textbf{0.417} & 94.62 & \textbf{67.31} & 72.00 & \textbf{4.95} \\
    w/o response-specific VTP & \xmark & \cmark & 7.53 & 7.06 & \textbf{0.417} & 93.58 & 67.26 & \textbf{81.50} & 4.65 \\
    w/o personalized ref. speech & \cmark & \xmark & \textbf{8.11} & \textbf{7.45} & 0.015 & \textbf{94.76} & 67.22 & 70.50 & 4.87 \\
    w/o both & \xmark & \xmark & \textbf{8.11} & \textbf{7.45} & 0.015 & 93.64 & 67.16 & 72.50 & 4.65 \\
    \bottomrule
  \end{tabularx}
\end{table*}

\begin{table}[htbp]
\centering

\begin{minipage}[t][3.2cm][t]{0.49\textwidth}
\centering

\caption{Effect of VTP supervision on speech QA. Both settings use
the same response-generation evaluation protocol.}
\label{tab:vtp_speech_qa_ablation}

\vfill

\scriptsize
\setlength{\tabcolsep}{5.0pt}
\renewcommand{\arraystretch}{1.8}
\begin{tabularx}{\linewidth}{
    @{}>{\raggedright\arraybackslash}X c c@{}
}
\toprule
\textbf{LLM Training Target}
& \textbf{CommonEval} $\uparrow$
& \textbf{BBH} $\uparrow$ \\
\midrule

Response only (w/o VTP)
& \textbf{3.82}
& \textbf{61.10} \\

VTP + response (ours)
& 3.71
& 58.70 \\
\bottomrule
\end{tabularx}

\end{minipage}
\hfill
\begin{minipage}[t][3.2cm][t]{0.49\textwidth}
\centering

\caption{Controlled upstream-source comparison.}
\label{tab:main_controlled_upstream}

\vfill

\scriptsize
\setlength{\tabcolsep}{1.5pt}
\renewcommand{\arraystretch}{1.5}
\begin{tabularx}{\linewidth}{
    @{}>{\raggedright\arraybackslash}X
    >{\raggedright\arraybackslash}X
    c c c@{}
}
\toprule
\textbf{Downstream}
& \textbf{Upstream}
& \textbf{SC} $\uparrow$
& \textbf{IQ} $\uparrow$
& \textbf{Sync-C} $\uparrow$ \\
\midrule

\multirow{2}{*}{OmniAvatar-1.3B}
& Qwen2.5-Omni
& 97.74
& 66.65
& \textbf{5.64} \\

& Ex-Omni-2D
& \textbf{98.19}
& \textbf{66.89}
& 5.48 \\
\midrule

\multirow{2}{*}{UniAVGen}
& Qwen2.5-Omni
& \textbf{96.07}
& \textbf{68.02}
& 4.15 \\

& Ex-Omni-2D
& 95.92
& 65.75
& \textbf{5.07} \\
\bottomrule
\end{tabularx}

\end{minipage}

\end{table}

\subsection{Effect of VTP Supervision.}
We isolate the effect of training the LLM to emit the structured VTP before its user-facing response. As shown in Table~\ref{tab:vtp_speech_qa_ablation}, response-only training without VTP supervision obtains 3.82 on CommonEval and 61.10 on BBH. Adding VTP supervision changes these scores to 3.71 and 58.70, corresponding to decreases of 0.11 and 2.40 points, respectively. The shared autoregressive channel gains an explicit visual-planning interface at the cost of lower speech-QA and reasoning scores.

\subsection{Effect of the Upstream LLM.}
As shown in Table~\ref{tab:main_controlled_upstream}, replacing
Qwen2.5-Omni with Ex-Omni-2D as the upstream model, while keeping the
downstream renderer and reference conditions fixed, produces mixed and renderer-dependent changes. For OmniAvatar, Ex-Omni-2D slightly improves SC and IQ but does not improve Sync-C. For UniAVGen, it substantially improves Sync-C, while SC and IQ decrease. Moreover, OmniAvatar retains the highest Sync-C among the tested renderers under either upstream source. These results show that the upstream LLM and its generated speech can affect video generation, but changing the LLM alone does not yield consistent improvements
across renderers and metrics. Our advantage instead stems from the native coupling of response-aware VTP semantics and frame-aligned speech units with a video generator specifically trained to consume them.

\begin{figure}[ht]
\begin{center}
    \includegraphics[width=0.95\linewidth]{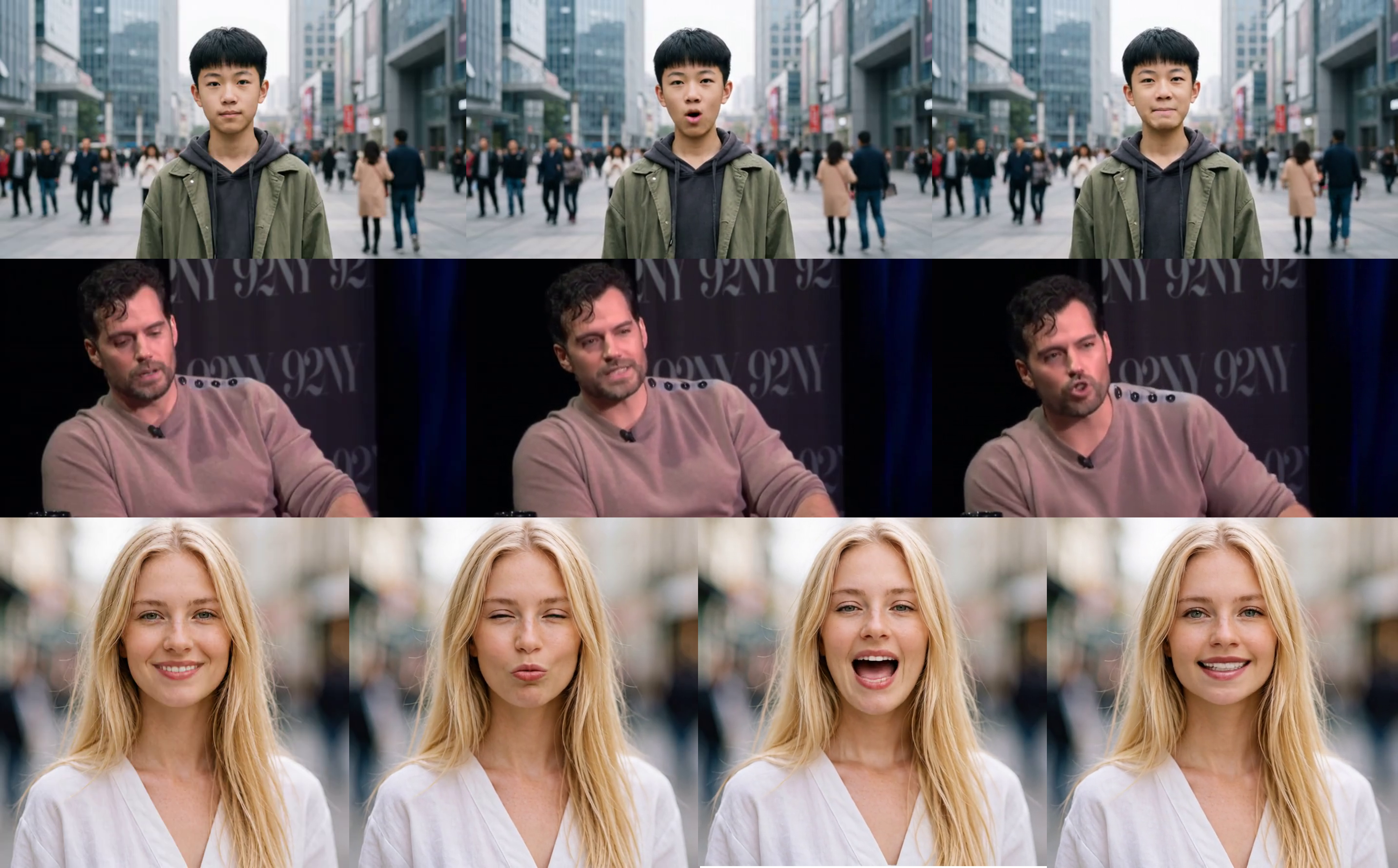}
    \textbf{(a)}\\[-0.4ex]
    \vspace{2pt}
    \includegraphics[width=0.95\linewidth]{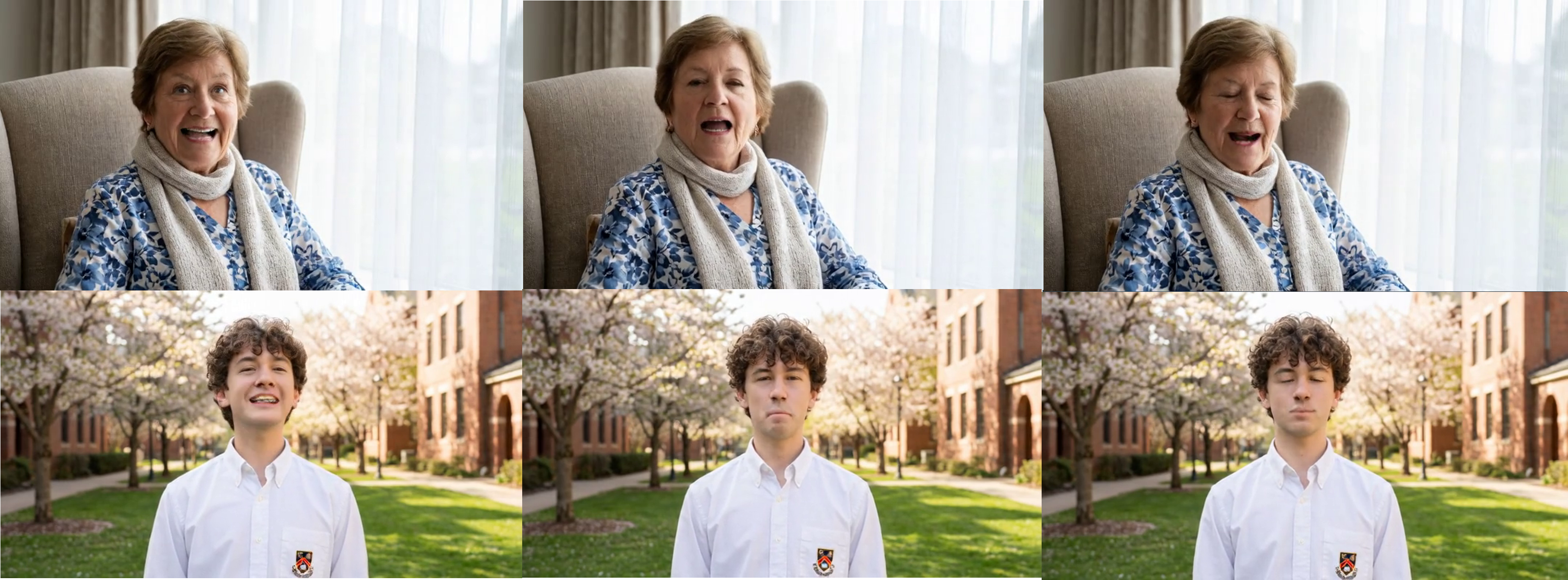}
    \textbf{(b)}\\[-0.4ex]
\end{center}
    \caption{Qualitative visualization of Ex-Omni-2D. \textbf{(a)} General VTP-guided omni-modal responses with head motion, facial expressions, and gestures. \textbf{(b)} Affect realization in dialogue-generated video responses with explicit emotion descriptions in the model-generated VTP.}
\label{fig:vis_results}
\end{figure}

\subsection{Visualization Analysis}
Figure~\ref{fig:vis_results} presents generated examples. Panel~(a) demonstrates that Ex-Omni-2D can generate visually coherent and appealing videos across different dialogue scenarios. In Panel~(b), we manually modify the \texttt{emotion} field of the VTP to examine emotion controllability. These edits produce visible expression changes in some cases, but the requested emotion may also appear weak or fail to emerge. This shows that the VTP acts as high-level semantic guidance rather than a strict control signal. The final motion may be jointly determined by the VTP, frame-aligned speech condition, and text/audio CFG scales.

\section{Conclusion}

Ex-Omni-2D turns a spoken omni-modal reply into a visible response by asking the dialogue model to produce two compact signals: a structured visual plan and multi-codebook speech units. The plan gives the avatar generator dialogue-specific scene and motion cues, while the units provide the timing shared by speech and video. This interface lets the speech, dialogue, and avatar pathways use the supervision available to each one, then reconnects them at inference. Our experiments show a clear split between the full-sequence Teacher, which gives the best visual quality, and the Prefix-Streaming Student, which starts playback earlier while trading off fidelity and synchronization.

\section*{Limitations}
Multi-codebook speech units provide a shared speech--video interface, but are currently harder to optimize for video conditioning than waveform-derived features. Due to limited computational resources, we have not verified whether larger-scale training can close this performance gap. Moreover, the Streaming
Student still exhibits drift over long sequences. The one-latent prefix partially mitigates this issue but does not fully eliminate accumulated errors.

\bibliographystyle{assets/plainnat}
\bibliography{citation}

@article{DBLP:journals/corr/abs-2503-20215,
  author       = {Jin Xu and
                  Zhifang Guo and
                  Jinzheng He and
                  Hangrui Hu and
                  Ting He and
                  Shuai Bai and
                  Keqin Chen and
                  Jialin Wang and
                  Yang Fan and
                  Kai Dang and
                  Bin Zhang and
                  Xiong Wang and
                  Yunfei Chu and
                  Junyang Lin},
  title        = {Qwen2.5-Omni Technical Report},
  journal      = {CoRR},
  volume       = {abs/2503.20215},
  year         = {2025}
}

@inproceedings{DBLP:conf/iclr/FangGZMZ025,
  author       = {Qingkai Fang and
                  Shoutao Guo and
                  Yan Zhou and
                  Zhengrui Ma and
                  Shaolei Zhang and
                  Yang Feng},
  title        = {LLaMA-Omni: Seamless Speech Interaction with Large Language Models},
  booktitle    = {The Thirteenth International Conference on Learning Representations},
  publisher    = {OpenReview.net},
  year         = {2025}
}

@inproceedings{DBLP:conf/acl/FangZGZ025,
  author       = {Qingkai Fang and
                  Yan Zhou and
                  Shoutao Guo and
                  Shaolei Zhang and
                  Yang Feng},
  title        = {LLaMA-Omni 2: LLM-based Real-time Spoken Chatbot with Autoregressive
                  Streaming Speech Synthesis},
  booktitle    = {Proceedings of the 63rd Annual Meeting of the Association for Computational Linguistics},
  pages        = {18617--18629},
  publisher    = {Association for Computational Linguistics},
  year         = {2025}
}

@article{DBLP:journals/corr/abs-2408-16725,
  author       = {Zhifei Xie and
                  Changqiao Wu},
  title        = {Mini-Omni: Language Models Can Hear, Talk While Thinking in Streaming},
  journal      = {CoRR},
  volume       = {abs/2408.16725},
  year         = {2024},
  eprinttype    = {arXiv},
  eprint       = {2408.16725}
}

@article{DBLP:journals/corr/abs-2512-19546,
  author       = {Ziqiao Peng and
                  Yi Chen and
                  Yifeng Ma and
                  Guozhen Zhang and
                  Zhiyao Sun and
                  Zixiang Zhou and
                  Youliang Zhang and
                  Zhengguang Zhou and
                  Zhaoxin Fan and
                  Hongyan Liu and
                  Yuan Zhou and
                  Qinglin Lu and
                  Jun He},
  title        = {ActAvatar: Temporally-Aware Precise Action Control for Talking Avatars},
  journal      = {CoRR},
  volume       = {abs/2512.19546},
  year         = {2025}
}

@article{DBLP:journals/corr/abs-2508-18621,
  author       = {Xin Gao and
                  Li Hu and
                  Siqi Hu and
                  Mingyang Huang and
                  Chaonan Ji and
                  Dechao Meng and
                  Jinwei Qi and
                  Penchong Qiao and
                  Zhen Shen and
                  Yafei Song and
                  Ke Sun and
                  Linrui Tian and
                  Guangyuan Wang and
                  Qi Wang and
                  Zhongjian Wang and
                  Jiayu Xiao and
                  Sheng Xu and
                  Bang Zhang and
                  Peng Zhang and
                  Xindi Zhang and
                  Zhe Zhang and
                  Jingren Zhou and
                  Lian Zhuo},
  title        = {Wan-S2V: Audio-Driven Cinematic Video Generation},
  journal      = {CoRR},
  volume       = {abs/2508.18621},
  year         = {2025}
}

@article{DBLP:journals/corr/abs-2506-18866,
  author       = {Qijun Gan and
                  Ruizi Yang and
                  Jianke Zhu and
                  Shaofei Xue and
                  Steven Hoi},
  title        = {OmniAvatar: Efficient Audio-Driven Avatar Video Generation with Adaptive
                  Body Animation},
  journal      = {CoRR},
  volume       = {abs/2506.18866},
  year         = {2025}
}

@article{DBLP:journals/corr/abs-2508-08248,
  author       = {Shuyuan Tu and
                  Yueming Pan and
                  Yinming Huang and
                  Xintong Han and
                  Zhen Xing and
                  Qi Dai and
                  Chong Luo and
                  Zuxuan Wu and
                  Yu{-}Gang Jiang},
  title        = {StableAvatar: Infinite-Length Audio-Driven Avatar Video Generation},
  journal      = {CoRR},
  volume       = {abs/2508.08248},
  year         = {2025}
}

@article{DBLP:journals/corr/abs-2505-20156,
  author       = {Yi Chen and
                  Sen Liang and
                  Zixiang Zhou and
                  Ziyao Huang and
                  Yifeng Ma and
                  Junshu Tang and
                  Qin Lin and
                  Yuan Zhou and
                  Qinglin Lu},
  title        = {HunyuanVideo-Avatar: High-Fidelity Audio-Driven Human Animation for
                  Multiple Characters},
  journal      = {CoRR},
  volume       = {abs/2505.20156},
  year         = {2025}
}

@inproceedings{DBLP:conf/cvpr/RuanMYH0FYJG23,
  author       = {Ludan Ruan and
                  Yiyang Ma and
                  Huan Yang and
                  Huiguo He and
                  Bei Liu and
                  Jianlong Fu and
                  Nicholas Jing Yuan and
                  Qin Jin and
                  Baining Guo},
  title        = {MM-Diffusion: Learning Multi-Modal Diffusion Models for Joint Audio
                  and Video Generation},
  booktitle    = {{IEEE/CVF} Conference on Computer Vision and Pattern Recognition,
                  {CVPR} 2023, Vancouver, BC, Canada, June 17-24, 2023},
  pages        = {10219--10228},
  publisher    = {{IEEE}},
  year         = {2023}
}

@inproceedings{liu2025javisdit,
  title       = {JavisDiT: Joint Audio-Video Diffusion Transformer with Hierarchical Spatio-Temporal Prior Synchronization}, 
  author      = {Liu, Kai and Li, Wei and Chen, Lai and Wu, Shengqiong and Zheng, Yanhao and Ji, Jiayi and Zhou, Fan and Luo, Jiebo and Liu, Ziwei and Fei, Hao and Chua, Tat-Seng},
  conference  = {The Fourteenth International Conference on Learning Representations},
  year        = {2026},
}

@article{Universe_1,
  author       = {Duomin Wang and
                  Wei Zuo and
                  Aojie Li and
                  Ling-Hao Chen and
                  Xinyao Liao and
                  Deyu Zhou and
                  Zixin Yin and
                  Xili Dai and
                  Daxin Jiang and
                  Gang Yu},
  title        = {UniVerse-1: Unified Audio-Video Generation via Stitching of Experts},
  journal      = {CoRR},
  volume       = {abs/2509.06155},
  year         = {2025}
}

@article{DBLP:journals/corr/abs-2410-11190,
  author       = {Zhifei Xie and
                  Changqiao Wu},
  title        = {Mini-Omni2: Towards Open-source GPT-4o with Vision, Speech and Duplex
                  Capabilities},
  journal      = {CoRR},
  volume       = {abs/2410.11190},
  year         = {2024},
  eprinttype    = {arXiv},
  eprint       = {2410.11190},
}

@inproceedings{luo2025openomni,
title={OpenOmni: Advancing Open-Source Omnimodal Large Language Models with Progressive Multimodal Alignment and Real-time Emotional Speech Synthesis},
author={Run Luo and Ting-En Lin and Haonan Zhang and Yuchuan Wu and Xiong Liu and Yongbin Li and Longze Chen and Jiaming Li and Lei Zhang and Xiaobo Xia and Hamid Alinejad-Rokny and Fei Huang and Min Yang},
booktitle={The Thirty-ninth Annual Conference on Neural Information Processing Systems},
year={2025}
}

@article{DBLP:journals/corr/abs-2506-09344,
  author       = {Inclusion AI and
                  Biao Gong and
                  Cheng Zou and
                  Chuanyang Zheng and
                  Chunluan Zhou and
                  Canxiang Yan and
                  Chunxiang Jin and
                  Chunjie Shen and
                  Dandan Zheng and
                  Fudong Wang and
                  Furong Xu and
                  Guangming Yao and
                  Jun Zhou and
                  Jingdong Chen and
                  Jianxin Sun and
                  Jiajia Liu and
                  Jianjiang Zhu and
                  Jun Peng and
                  Kaixiang Ji and
                  Kaiyou Song and
                  Kaimeng Ren and
                  Libin Wang and
                  Lixiang Ru and
                  Lele Xie and
                  Longhua Tan and
                  Lyuxin Xue and
                  Lan Wang and
                  Mochen Bai and
                  Ning Gao and
                  Pei Chen and
                  Qingpei Guo and
                  Qinglong Zhang and
                  Qiang Xu and
                  Rui Liu and
                  Ruijie Xiong and
                  Sirui Gao and
                  Tinghao Liu and
                  Taisong Li and
                  Weilong Chai and
                  Xinyu Xiao and
                  Xiaomei Wang and
                  Xiaoxue Chen and
                  Xiao Lu and
                  Xiaoyu Li and
                  Xingning Dong and
                  Xuzheng Yu and
                  Yi Yuan and
                  Yuting Gao and
                  Yunxiao Sun and
                  Yipeng Chen and
                  Yifei Wu and
                  Yongjie Lyu and
                  Ziping Ma and
                  Zipeng Feng and
                  Zhijiang Fang and
                  Zhihao Qiu and
                  Ziyuan Huang and
                  Zhengyu He},
  title        = {Ming-Omni: {A} Unified Multimodal Model for Perception and Generation},
  journal      = {CoRR},
  volume       = {abs/2506.09344},
  year         = {2025}
}

@inproceedings{DBLP:conf/acl/AoWZ0RW0KLZWQ0W22,
  author       = {Junyi Ao and
                  Rui Wang and
                  Long Zhou and
                  Chengyi Wang and
                  Shuo Ren and
                  Yu Wu and
                  Shujie Liu and
                  Tom Ko and
                  Qing Li and
                  Yu Zhang and
                  Zhihua Wei and
                  Yao Qian and
                  Jinyu Li and
                  Furu Wei},
  title        = {SpeechT5: Unified-Modal Encoder-Decoder Pre-Training for Spoken Language
                  Processing},
  booktitle    = {Proceedings of the 60th Annual Meeting of the Association for Computational Linguistics},
  pages        = {5723--5738},
  year         = {2022}
}

@article{DBLP:journals/corr/abs-2401-13527,
  author       = {Dong Zhang and
                  Xin Zhang and
                  Jun Zhan and
                  Shimin Li and
                  Yaqian Zhou and
                  Xipeng Qiu},
  title        = {SpeechGPT-Gen: Scaling Chain-of-Information Speech Generation},
  journal      = {CoRR},
  volume       = {abs/2401.13527},
  year         = {2024},
  doi          = {10.48550/ARXIV.2401.13527}
}

@article{DBLP:journals/corr/abs-2410-00037,
  author       = {Alexandre D{\'{e}}fossez and
                  Laurent Mazar{\'{e}} and
                  Manu Orsini and
                  Am{\'{e}}lie Royer and
                  Patrick P{\'{e}}rez and
                  Herv{\'{e}} J{\'{e}}gou and
                  Edouard Grave and
                  Neil Zeghidour},
  title        = {Moshi: a speech-text foundation model for real-time dialogue},
  journal      = {CoRR},
  volume       = {abs/2410.00037},
  year         = {2024}
}

@article{DBLP:journals/corr/abs-2412-02612,
  author       = {Aohan Zeng and
                  Zhengxiao Du and
                  Mingdao Liu and
                  Kedong Wang and
                  Shengmin Jiang and
                  Lei Zhao and
                  Yuxiao Dong and
                  Jie Tang},
  title        = {GLM-4-Voice: Towards Intelligent and Human-Like End-to-End Spoken
                  Chatbot},
  journal      = {CoRR},
  volume       = {abs/2412.02612},
  year         = {2024}
}

@article{DBLP:journals/corr/abs-2511-00279,
  author       = {Meituan LongCat Team},
  title        = {LongCat-Flash-Omni Technical Report},
  journal      = {CoRR},
  volume       = {abs/2511.00279},
  year         = {2025},
}

@InProceedings{Zhang_2023_CVPR,
    author    = {Zhang, Wenxuan and Cun, Xiaodong and Wang, Xuan and Zhang, Yong and Shen, Xi and Guo, Yu and Shan, Ying and Wang, Fei},
    title     = {SadTalker: Learning Realistic 3D Motion Coefficients for Stylized Audio-Driven Single Image Talking Face Animation},
    booktitle = {Proceedings of the IEEE/CVF Conference on Computer Vision and Pattern Recognition (CVPR)},
    month     = {June},
    year      = {2023},
    pages     = {8652-8661}
}

@misc{xu2024hallohierarchicalaudiodrivenvisual,
      title={Hallo: Hierarchical Audio-Driven Visual Synthesis for Portrait Image Animation}, 
      author={Mingwang Xu and Hui Li and Qingkun Su and Hanlin Shang and Liwei Zhang and Ce Liu and Jingdong Wang and Yao Yao and Siyu Zhu},
      year={2024},
      eprint={2406.08801},
      archivePrefix={arXiv},
      primaryClass={cs.CV} 
}

@inproceedings{10.1145/3394171.3413532,
author = {Prajwal, K R and Mukhopadhyay, Rudrabha and Namboodiri, Vinay P. and Jawahar, C.V.},
title = {A Lip Sync Expert Is All You Need for Speech to Lip Generation In the Wild},
year = {2020},
isbn = {9781450379885},
booktitle = {Proceedings of the 28th ACM International Conference on Multimedia},
pages = {484–492},
numpages = {9},
}

@inproceedings{NEURIPS2023_98c50f47,
 author = {Luo, Simian and Yan, Chuanhao and Hu, Chenxu and Zhao, Hang},
 booktitle = {Advances in Neural Information Processing Systems},
 editor = {A. Oh and T. Naumann and A. Globerson and K. Saenko and M. Hardt and S. Levine},
 pages = {48855--48876},
 publisher = {Curran Associates, Inc.},
 title = {Diff-Foley: Synchronized Video-to-Audio Synthesis with Latent Diffusion Models},
 volume = {36},
 year = {2023}
}

@InProceedings{Cheng_2025_CVPR,
    author    = {Cheng, Ho Kei and Ishii, Masato and Hayakawa, Akio and Shibuya, Takashi and Schwing, Alexander and Mitsufuji, Yuki},
    title     = {MMAudio: Taming Multimodal Joint Training for High-Quality Video-to-Audio Synthesis},
    booktitle = {Proceedings of the IEEE/CVF Conference on Computer Vision and Pattern Recognition (CVPR)},
    month     = {June},
    year      = {2025},
    pages     = {28901-28911}
}

@InProceedings{Sung-Bin_2025_ICCV,
    author    = {Sung-Bin, Kim and Choi, Jeongsoo and Peng, Puyuan and Chung, Joon Son and Oh, Tae-Hyun and Harwath, David},
    title     = {VoiceCraft-Dub: Automated Video Dubbing with Neural Codec Language Models},
    booktitle = {Proceedings of the IEEE/CVF International Conference on Computer Vision (ICCV)},
    month     = {October},
    year      = {2025},
    pages     = {14623-14632}
}

@misc{liu2024syncflowtemporallyalignedjoint,
      title={SyncFlow: Toward Temporally Aligned Joint Audio-Video Generation from Text}, 
      author={Haohe Liu and Gael Le Lan and Xinhao Mei and Zhaoheng Ni and Anurag Kumar and Varun Nagaraja and Wenwu Wang and Mark D. Plumbley and Yangyang Shi and Vikas Chandra},
      year={2024},
      eprint={2412.15220}
}

@misc{zhao2025uniformunifiedmultitaskdiffusion,
      title={UniForm: A Unified Multi-Task Diffusion Transformer for Audio-Video Generation}, 
      author={Lei Zhao and Linfeng Feng and Dongxu Ge and Rujin Chen and Fangqiu Yi and Chi Zhang and Xiao-Lei Zhang and Xuelong Li},
      year={2025},
      eprint={2502.03897},
}

@misc{zhang2026uniavgenunifiedaudiovideo,
      title={UniAVGen: Unified Audio and Video Generation with Asymmetric Cross-Modal Interactions}, 
      author={Guozhen Zhang and Zixiang Zhou and Teng Hu and Ziqiao Peng and Youliang Zhang and Yi Chen and Yuan Zhou and Qinglin Lu and Limin Wang},
      year={2026},
      eprint={2511.03334},
      archivePrefix={arXiv},
      primaryClass={cs.CV}
}

@misc{low2025ovitwinbackbonecrossmodal,
      title={Ovi: Twin Backbone Cross-Modal Fusion for Audio-Video Generation}, 
      author={Chetwin Low and Weimin Wang and Calder Katyal},
      year={2025},
      eprint={2510.01284}
}

@INPROCEEDINGS{7178964,
  author={Panayotov, Vassil and Chen, Guoguo and Povey, Daniel and Khudanpur, Sanjeev},
  booktitle={2015 IEEE International Conference on Acoustics, Speech and Signal Processing (ICASSP)}, 
  title={Librispeech: An ASR corpus based on public domain audio books}, 
  year={2015},
  volume={},
  number={},
  pages={5206-5210}
}

@INPROCEEDINGS{10832365,
  author={He, Haorui and Shang, Zengqiang and Wang, Chaoren and Li, Xuyuan and Gu, Yicheng and Hua, Hua and Liu, Liwei and Yang, Chen and Li, Jiaqi and Shi, Peiyang and Wang, Yuancheng and Chen, Kai and Zhang, Pengyuan and Wu, Zhizheng},
  booktitle={2024 IEEE Spoken Language Technology Workshop (SLT)}, 
  title={Emilia: An Extensive, Multilingual, and Diverse Speech Dataset For Large-Scale Speech Generation}, 
  year={2024},
  volume={},
  number={},
  pages={885-890}
}

@misc{hu2026qwen3ttstechnicalreport,
      title={Qwen3-TTS Technical Report}, 
      author={Hangrui Hu and Xinfa Zhu and Ting He and Dake Guo and Bin Zhang and Xiong Wang and Zhifang Guo and Ziyue Jiang and Hongkun Hao and Zishan Guo and Xinyu Zhang and Pei Zhang and Baosong Yang and Jin Xu and Jingren Zhou and Junyang Lin},
      year={2026},
      eprint={2601.15621}
}

@article{zhang2025speakervid,
  title={Speakervid-5m: A large-scale high-quality dataset for audio-visual dyadic interactive human generation},
  author={Zhang, Youliang and Li, Zhaoyang and Wang, Duomin and Zhang, Jiahe and Zhou, Deyu and Yin, Zixin and Dai, Xili and Yu, Gang and Li, Xiu},
  journal={arXiv preprint arXiv:2507.09862},
  year={2025}
}

@article{chen2024voicebenchbenchmarkingllmbasedvoice,
  title={VoiceBench: Benchmarking LLM-Based Voice Assistants},
  author={Chen, Yiming and Yue, Xianghu and Zhang, Chen and Gao, Xiaoxue and Tan, Robby T. and Li, Haizhou},
  journal={arXiv preprint arXiv:2410.17196},
  year={2024}
}

@misc{chen2024echomimic,
  title={EchoMimic: Lifelike Audio-Driven Portrait Animations through Editable Landmark Conditioning},
  author={Zhiyuan Chen and Jiajiong Cao and Zhiquan Chen and Yuming Li and Chenguang Ma},
  year={2024},
  eprint={2407.08136},
  archivePrefix={arXiv},
  primaryClass={cs.CV}
}

@article{tjandra2025aes,
    title={Meta Audiobox Aesthetics: Unified Automatic Quality Assessment for Speech, Music, and Sound},
    author={Andros Tjandra and Yi-Chiao Wu and Baishan Guo and John Hoffman and Brian Ellis and Apoorv Vyas and Bowen Shi and Sanyuan Chen and Matt Le and Nick Zacharov and Carleigh Wood and Ann Lee and Wei-Ning Hsu},
    year={2025},
    journal={arXiv preprint arXiv:2502.05139}
}

@InProceedings{huang2023vbench,
 title={{VBench}: Comprehensive Benchmark Suite for Video Generative Models},
 author={Huang, Ziqi and He, Yinan and Yu, Jiashuo and Zhang, Fan and Si, Chenyang and Jiang, Yuming and Zhang, Yuanhan and Wu, Tianxing and Jin, Qingyang and Chanpaisit, Nattapol and Wang, Yaohui and Chen, Xinyuan and Wang, Limin and Lin, Dahua and Qiao, Yu and Liu, Ziwei},
 booktitle={Proceedings of the IEEE/CVF Conference on Computer Vision and Pattern Recognition},
 year={2024}
}

@article{wan2025,
      title={Wan: Open and Advanced Large-Scale Video Generative Models}, 
      author={Team Wan and Ang Wang and Baole Ai and Bin Wen and Chaojie Mao and Chen-Wei Xie and Di Chen and Feiwu Yu and Haiming Zhao and Jianxiao Yang and Jianyuan Zeng and Jiayu Wang and Jingfeng Zhang and Jingren Zhou and Jinkai Wang and Jixuan Chen and Kai Zhu and Kang Zhao and Keyu Yan and Lianghua Huang and Mengyang Feng and Ningyi Zhang and Pandeng Li and Pingyu Wu and Ruihang Chu and Ruili Feng and Shiwei Zhang and Siyang Sun and Tao Fang and Tianxing Wang and Tianyi Gui and Tingyu Weng and Tong Shen and Wei Lin and Wei Wang and Wei Wang and Wenmeng Zhou and Wente Wang and Wenting Shen and Wenyuan Yu and Xianzhong Shi and Xiaoming Huang and Xin Xu and Yan Kou and Yangyu Lv and Yifei Li and Yijing Liu and Yiming Wang and Yingya Zhang and Yitong Huang and Yong Li and You Wu and Yu Liu and Yulin Pan and Yun Zheng and Yuntao Hong and Yupeng Shi and Yutong Feng and Zeyinzi Jiang and Zhen Han and Zhi-Fan Wu and Ziyu Liu},
      journal = {arXiv preprint arXiv:2503.20314},
      year={2025}
}

@article{yang2025qwen3,
  title={Qwen3 technical report},
  author={Yang, An and Li, Anfeng and Yang, Baosong and Zhang, Beichen and Hui, Binyuan and Zheng, Bo and Yu, Bowen and Gao, Chang and Huang, Chengen and Lv, Chenxu and others},
  journal={arXiv preprint arXiv:2505.09388},
  year={2025}
}

@article{fu2026vita,
  title={Vita-1.5: Towards gpt-4o level real-time vision and speech interaction},
  author={Fu, Chaoyou and Lin, Haojia and Wang, Xiong and Shen, Yunhang and Liu, Xiaoyu and Cao, Haoyu and Long, Zuwei and Gao, Heting and Li, Ke and MA, Long and others},
  journal={Advances in Neural Information Processing Systems},
  volume={38},
  pages={75300--75320},
  year={2026}
}

@inproceedings{chen2025slam,
  title={Slam-omni: Timbre-controllable voice interaction system with single-stage training},
  author={Chen, Wenxi and Ma, Ziyang and Yan, Ruiqi and Liang, Yuzhe and Li, Xiquan and Xu, Ruiyang and Niu, Zhikang and Zhu, Yanqiao and Yang, Yifan and Liu, Zhanxun and others},
  booktitle={Findings of the Association for Computational Linguistics: ACL 2025},
  pages={2262--2282},
  year={2025}
}

@article{gu2026anyflow,
  title={Anyflow: Any-step video diffusion model with on-policy flow map distillation},
  author={Gu, Yuchao and Fang, Guian and Jiang, Yuxin and Mao, Weijia and Han, Song and Cai, Han and Shou, Mike Zheng},
  journal={arXiv preprint arXiv:2605.13724},
  year={2026}
}

@article{bai2025qwen3,
  title={Qwen3-vl technical report},
  author={Bai, Shuai and Cai, Yuxuan and Chen, Ruizhe and Chen, Keqin and Chen, Xionghui and Cheng, Zesen and Deng, Lianghao and Ding, Wei and Gao, Chang and Ge, Chunjiang and others},
  journal={arXiv preprint arXiv:2511.21631},
  year={2025}
}

@InProceedings{Sun_2026_CVPR,
    author    = {Sun, Zhiyao and Peng, Ziqiao and Ma, Yifeng and Chen, Yi and Zhou, Zhengguang and Zhou, Zixiang and Zhang, Guozhen and Zhang, Youliang and Zhou, Yuan and Lu, Qinglin and Liu, Yong-Jin},
    title     = {StreamAvatar: Streaming Diffusion Models for Real-Time Interactive Human Avatars},
    booktitle = {Proceedings of the IEEE/CVF Conference on Computer Vision and Pattern Recognition (CVPR)},
    month     = {June},
    year      = {2026},
    pages     = {10887-10897}
}

@inproceedings{zhang-etal-2025-omnicharacter,
  title={OmniCharacter: Towards immersive role-playing agents with seamless speech-language personality interaction},
  author={Zhang, Haonan and Luo, Run and Liu, Xiong and Wu, Yuchuan and Lin, Ting-En and Zeng, Pengpeng and Qu, Qiang and Fang, Feiteng and Yang, Min and Gao, Lianli and others},
  booktitle={Proceedings of the 63rd Annual Meeting of the Association for Computational Linguistics (Volume 1: Long Papers)},
  pages={26318--26331},
  year={2025}
}

@inproceedings{wang2025fantasytalking,
  title={Fantasytalking: Realistic talking portrait generation via coherent motion synthesis},
  author={Wang, Mengchao and Wang, Qiang and Jiang, Fan and Fan, Yaqi and Zhang, Yunpeng and Qi, Yonggang and Zhao, Kun and Xu, Mu},
  booktitle={Proceedings of the 33rd ACM International Conference on Multimedia},
  pages={9891--9900},
  year={2025}
}

@article{anastassiou2024seed,
  title={Seed-tts: A family of high-quality versatile speech generation models},
  author={Anastassiou, Philip and Chen, Jiawei and Chen, Jitong and Chen, Yuanzhe and Chen, Zhuo and Chen, Ziyi and Cong, Jian and Deng, Lelai and Ding, Chuang and Gao, Lu and others},
  journal={arXiv preprint arXiv:2406.02430},
  year={2024}
}

@article{Chen2021WavLM,
  author       = {Sanyuan Chen and
                  Chengyi Wang and
                  Zhengyang Chen and
                  Yu Wu and
                  Shujie Liu and
                  Zhuo Chen and
                  Jinyu Li and
                  Naoyuki Kanda and
                  Takuya Yoshioka and
                  Xiong Xiao and
                  Jian Wu and
                  Long Zhou and
                  Shuo Ren and
                  Yanmin Qian and
                  Yao Qian and
                  Jian Wu and
                  Michael Zeng and
                  Xiangzhan Yu and
                  Furu Wei},
  title        = {WavLM: Large-Scale Self-Supervised Pre-Training for Full Stack Speech
                  Processing},
  journal      = {{IEEE} J. Sel. Top. Signal Process.},
  volume       = {16},
  number       = {6},
  pages        = {1505--1518},
  year         = {2022}
}

@inproceedings{desplanques2020ecapa,
  title={{ECAPA-TDNN: Emphasized Channel Attention, propagation and aggregation in TDNN based speaker verification}},
  author={Desplanques, Brecht and Thienpondt, Jenthe and Demuynck, Kris},
  booktitle={Interspeech 2020},
  pages={3830--3834},
  year={2020}
}

@InProceedings{Chung16a,
  author       = "Chung, J.~S. and Zisserman, A.",
  title        = "Out of time: automated lip sync in the wild",
  booktitle    = "Workshop on Multi-view Lip-reading, ACCV",
  year         = "2016",
}

@misc{alpaca_eval,
  author = {Xuechen Li and Tianyi Zhang and Yann Dubois and Rohan Taori and Ishaan Gulrajani and Carlos Guestrin and Percy Liang and Tatsunori B. Hashimoto },
  title = {AlpacaEval: An Automatic Evaluator of Instruction-following Models},
  year = {2023},
  month = {5},
  publisher = {GitHub},
  journal = {GitHub repository},
  howpublished = {\url{https://github.com/tatsu-lab/alpaca_eval}}
}

@article{suzgun2022challenging,
  title={Challenging BIG-Bench Tasks and Whether Chain-of-Thought Can Solve Them},
  author={Suzgun, Mirac and Scales, Nathan and Sch{\"a}rli, Nathanael and Gehrmann, Sebastian and Tay, Yi and Chung, Hyung Won and Chowdhery, Aakanksha and Le, Quoc V and Chi, Ed H and Zhou, Denny and and Wei, Jason},
  journal={arXiv preprint arXiv:2210.09261},
  year={2022}
}

@inproceedings{caron2021emerging,
  title={Emerging Properties in Self-Supervised Vision Transformers},
  author={Caron, Mathilde and Touvron, Hugo and Misra, Ishan and J\'egou, Herv\'e  and Mairal, Julien and Bojanowski, Piotr and Joulin, Armand},
  booktitle={Proceedings of the International Conference on Computer Vision (ICCV)},
  year={2021}
}

@article{yolox2021,
title={YOLOX: Exceeding YOLO Series in 2021},
author={Ge, Zheng and Liu, Songtao and Wang, Feng and Li, Zeming and Sun, Jian},
journal={arXiv preprint arXiv:2107.08430},
year={2021}
}

@inproceedings{yang2023effective,
  title={Effective whole-body pose estimation with two-stages distillation},
  author={Yang, Zhendong and Zeng, Ailing and Yuan, Chun and Li, Yu},
  booktitle={Proceedings of the IEEE/CVF International Conference on Computer Vision},
  pages={4210--4220},
  year={2023}
}

@misc{zhang2017s3fdsingleshotscaleinvariant,
      title={S$^3$FD: Single Shot Scale-invariant Face Detector}, 
      author={Shifeng Zhang and Xiangyu Zhu and Zhen Lei and Hailin Shi and Xiaobo Wang and Stan Z. Li},
      year={2017},
      eprint={1708.05237}
}

@inproceedings{wu2023dover,
      title={Exploring Video Quality Assessment on User Generated Contents from Aesthetic and Technical Perspectives}, 
      author={Wu, Haoning and Zhang, Erli and Liao, Liang and Chen, Chaofeng and Hou, Jingwen Hou and Wang, Annan and Sun, Wenxiu Sun and Yan, Qiong and Lin, Weisi},
      year={2023},
      booktitle={International Conference on Computer Vision (ICCV)},
}

@InProceedings{Su_2020_CVPR,
author = {Su, Shaolin and Yan, Qingsen and Zhu, Yu and Zhang, Cheng and Ge, Xin and Sun, Jinqiu and Zhang, Yanning},
title = {Blindly Assess Image Quality in the Wild Guided by a Self-Adaptive Hyper Network},
booktitle = {IEEE/CVF Conference on Computer Vision and Pattern Recognition (CVPR)},
month = {June},
year = {2020}
}

@book{opencv,
author = {Bradski, Gary},
year = {2000},
month = {11},
pages = {},
title = {The Opencv Library},
volume = {25},
journal = {Dr. Dobb's J. Softw. Tools}
}

@software{castellano_pyscenedetect,
  author = {Castellano, Brandon},
  title = {PySceneDetect},
  url = {https://www.scenedetect.com},
  note = {Video Cut Detection and Analysis Tool}
}

@misc{lindevs_yolov8_face,
  author = {{Lindevs}},
  title = {YOLOv8-Face},
  howpublished = {\url{https://github.com/lindevs/yolov8-face}},
  note = {Accessed: 2026-08-11}
}

@misc{glm2024chatglm,
      title={ChatGLM: A Family of Large Language Models from GLM-130B to GLM-4 All Tools}, 
      author={Team GLM and Aohan Zeng and Bin Xu and Bowen Wang and Chenhui Zhang and Da Yin and Diego Rojas and Guanyu Feng and Hanlin Zhao and Hanyu Lai and Hao Yu and Hongning Wang and Jiadai Sun and Jiajie Zhang and Jiale Cheng and Jiayi Gui and Jie Tang and Jing Zhang and Juanzi Li and Lei Zhao and Lindong Wu and Lucen Zhong and Mingdao Liu and Minlie Huang and Peng Zhang and Qinkai Zheng and Rui Lu and Shuaiqi Duan and Shudan Zhang and Shulin Cao and Shuxun Yang and Weng Lam Tam and Wenyi Zhao and Xiao Liu and Xiao Xia and Xiaohan Zhang and Xiaotao Gu and Xin Lv and Xinghan Liu and Xinyi Liu and Xinyue Yang and Xixuan Song and Xunkai Zhang and Yifan An and Yifan Xu and Yilin Niu and Yuantao Yang and Yueyan Li and Yushi Bai and Yuxiao Dong and Zehan Qi and Zhaoyu Wang and Zhen Yang and Zhengxiao Du and Zhenyu Hou and Zihan Wang},
      year={2024},
      eprint={2406.12793},
      archivePrefix={arXiv}
}

@article{baichuan2023baichuan2,
  title={Baichuan 2: Open Large-scale Language Models},
  author={Baichuan},
  journal={arXiv preprint arXiv:2309.10305},
  year={2023}
}

@misc{cai2024internlm2,
      title={InternLM2 Technical Report},
      author={Zheng Cai and Maosong Cao and Haojiong Chen and Kai Chen and Keyu Chen and Xin Chen and Xun Chen and Zehui Chen and Zhi Chen and Pei Chu and Xiaoyi Dong and Haodong Duan and Qi Fan and Zhaoye Fei and Yang Gao and Jiaye Ge and Chenya Gu and Yuzhe Gu and Tao Gui and Aijia Guo and Qipeng Guo and Conghui He and Yingfan Hu and Ting Huang and Tao Jiang and Penglong Jiao and Zhenjiang Jin and Zhikai Lei and Jiaxing Li and Jingwen Li and Linyang Li and Shuaibin Li and Wei Li and Yining Li and Hongwei Liu and Jiangning Liu and Jiawei Hong and Kaiwen Liu and Kuikun Liu and Xiaoran Liu and Chengqi Lv and Haijun Lv and Kai Lv and Li Ma and Runyuan Ma and Zerun Ma and Wenchang Ning and Linke Ouyang and Jiantao Qiu and Yuan Qu and Fukai Shang and Yunfan Shao and Demin Song and Zifan Song and Zhihao Sui and Peng Sun and Yu Sun and Huanze Tang and Bin Wang and Guoteng Wang and Jiaqi Wang and Jiayu Wang and Rui Wang and Yudong Wang and Ziyi Wang and Xingjian Wei and Qizhen Weng and Fan Wu and Yingtong Xiong and Chao Xu and Ruiliang Xu and Hang Yan and Yirong Yan and Xiaogui Yang and Haochen Ye and Huaiyuan Ying and Jia Yu and Jing Yu and Yuhang Zang and Chuyu Zhang and Li Zhang and Pan Zhang and Peng Zhang and Ruijie Zhang and Shuo Zhang and Songyang Zhang and Wenjian Zhang and Wenwei Zhang and Xingcheng Zhang and Xinyue Zhang and Hui Zhao and Qian Zhao and Xiaomeng Zhao and Fengzhe Zhou and Zaida Zhou and Jingming Zhuo and Yicheng Zou and Xipeng Qiu and Yu Qiao and Dahua Lin},
      year={2024},
      eprint={2403.17297},
      archivePrefix={arXiv}
}

@article{zhou2023characterglm,
  title={CharacterGLM: Customizing Chinese Conversational AI Characters with Large Language Models},
  author={Zhou, Jinfeng and Chen, Zhuang and Wan, Dazhen and Wen, Bosi and Song, Yi and Yu, Jifan and Huang, Yongkang and Peng, Libiao and Yang, Jiaming and Xiao, Xiyao and others},
  journal={arXiv preprint arXiv:2311.16832},
  year={2023}
}

@article{grattafiori2024llama,
  title={The Llama 3 Herd of Models},
  author={Grattafiori, Aaron and Dubey, Abhimanyu and Jauhri, Abhinav and others},
  journal={arXiv preprint arXiv:2407.21783},
  year={2024}
}

@article{DBLP:journals/corr/abs-2309-16609,
  author       = {Jinze Bai and
                  Shuai Bai and
                  Yunfei Chu and
                  Zeyu Cui and
                  Kai Dang and
                  Xiaodong Deng and
                  Yang Fan and
                  Wenbin Ge and
                  Yu Han and
                  Fei Huang and
                  Binyuan Hui and
                  Luo Ji and
                  Mei Li and
                  Junyang Lin and
                  Runji Lin and
                  Dayiheng Liu and
                  Gao Liu and
                  Chengqiang Lu and
                  Keming Lu and
                  Jianxin Ma and
                  Rui Men and
                  Xingzhang Ren and
                  Xuancheng Ren and
                  Chuanqi Tan and
                  Sinan Tan and
                  Jianhong Tu and
                  Peng Wang and
                  Shijie Wang and
                  Wei Wang and
                  Shengguang Wu and
                  Benfeng Xu and
                  Jin Xu and
                  An Yang and
                  Hao Yang and
                  Jian Yang and
                  Shusheng Yang and
                  Yang Yao and
                  Bowen Yu and
                  Hongyi Yuan and
                  Zheng Yuan and
                  Jianwei Zhang and
                  Xingxuan Zhang and
                  Yichang Zhang and
                  Zhenru Zhang and
                  Chang Zhou and
                  Jingren Zhou and
                  Xiaohuan Zhou and
                  Tianhang Zhu},
  title        = {Qwen Technical Report},
  journal      = {CoRR},
  volume       = {abs/2309.16609},
  year         = {2023},
  doi          = {10.48550/ARXIV.2309.16609}
}

@article{DBLP:journals/corr/abs-2407-10671,
  author       = {An Yang and
                  Baosong Yang and
                  Binyuan Hui and
                  Bo Zheng and
                  Bowen Yu and
                  Chang Zhou and
                  Chengpeng Li and
                  Chengyuan Li and
                  Dayiheng Liu and
                  Fei Huang and
                  Guanting Dong and
                  Haoran Wei and
                  Huan Lin and
                  Jialong Tang and
                  Jialin Wang and
                  Jian Yang and
                  Jianhong Tu and
                  Jianwei Zhang and
                  Jianxin Ma and
                  Jianxin Yang and
                  Jin Xu and
                  Jingren Zhou and
                  Jinze Bai and
                  Jinzheng He and
                  Junyang Lin and
                  Kai Dang and
                  Keming Lu and
                  Keqin Chen and
                  Kexin Yang and
                  Mei Li and
                  Mingfeng Xue and
                  Na Ni and
                  Pei Zhang and
                  Peng Wang and
                  Ru Peng and
                  Rui Men and
                  Ruize Gao and
                  Runji Lin and
                  Shijie Wang and
                  Shuai Bai and
                  Sinan Tan and
                  Tianhang Zhu and
                  Tianhao Li and
                  Tianyu Liu and
                  Wenbin Ge and
                  Xiaodong Deng and
                  Xiaohuan Zhou and
                  Xingzhang Ren and
                  Xinyu Zhang and
                  Xipin Wei and
                  Xuancheng Ren and
                  Xuejing Liu and
                  Yang Fan and
                  Yang Yao and
                  Yichang Zhang and
                  Yu Wan and
                  Yunfei Chu and
                  Yuqiong Liu and
                  Zeyu Cui and
                  Zhenru Zhang and
                  Zhifang Guo and
                  Zhihao Fan},
  title        = {Qwen2 Technical Report},
  journal      = {CoRR},
  volume       = {abs/2407.10671},
  year         = {2024},
  doi          = {10.48550/ARXIV.2407.10671}
}

@article{DBLP:journals/corr/abs-2412-15115,
  author       = {An Yang and
                  Baosong Yang and
                  Beichen Zhang and
                  Binyuan Hui and
                  Bo Zheng and
                  Bowen Yu and
                  Chengyuan Li and
                  Dayiheng Liu and
                  Fei Huang and
                  Haoran Wei and
                  Huan Lin and
                  Jian Yang and
                  Jianhong Tu and
                  Jianwei Zhang and
                  Jianxin Yang and
                  Jiaxi Yang and
                  Jingren Zhou and
                  Junyang Lin and
                  Kai Dang and
                  Keming Lu and
                  Keqin Bao and
                  Kexin Yang and
                  Le Yu and
                  Mei Li and
                  Mingfeng Xue and
                  Pei Zhang and
                  Qin Zhu and
                  Rui Men and
                  Runji Lin and
                  Tianhao Li and
                  Tingyu Xia and
                  Xingzhang Ren and
                  Xuancheng Ren and
                  Yang Fan and
                  Yang Su and
                  Yichang Zhang and
                  Yu Wan and
                  Yuqiong Liu and
                  Zeyu Cui and
                  Zhenru Zhang and
                  Zihan Qiu},
  title        = {Qwen2.5 Technical Report},
  journal      = {CoRR},
  volume       = {abs/2412.15115},
  year         = {2024},
  doi          = {10.48550/ARXIV.2412.15115}
}

@misc{openai2024gpt4omini,
  author       = {{OpenAI}},
  title        = {GPT-4o mini: Advancing Cost-Efficient Intelligence},
  year         = {2024},
  month        = jul,
  howpublished = {\url{https://openai.com/index/gpt-4o-mini-advancing-cost-efficient-intelligence/}},
  note         = {Accessed: 2026-09-01}
}

@inproceedings{DBLP:conf/eccv/TeedD20,
  author       = {Zachary Teed and
                  Jia Deng},
  title        = {{RAFT:} Recurrent All-Pairs Field Transforms for Optical Flow},
  booktitle    = {Computer Vision - {ECCV} 2020 - 16th European Conference, Glasgow,
                  UK, August 23-28, 2020, Proceedings, Part {II}},
  volume       = {12347},
  pages        = {402--419},
  publisher    = {Springer},
  year         = {2020}
}

@misc{insightface,
  author       = {{InsightFace Contributors}},
  title        = {{InsightFace}: 2D and 3D Face Analysis Project},
  howpublished = {\url{https://github.com/deepinsight/insightface}},
  year         = {2026},
  note         = {Accessed: 2026-09-01}
}

@inproceedings{DBLP:conf/nips/BaevskiZMA20,
  author       = {Alexei Baevski and
                  Yuhao Zhou and
                  Abdelrahman Mohamed and
                  Michael Auli},
  title        = {wav2vec 2.0: {A} Framework for Self-Supervised Learning of Speech
                  Representations},
  booktitle    = {NeurIPS 2020},
  year         = {2020}
}

\clearpage
\appendix

\section{Method Details}
\paragraph{Streaming Student Method Details.}
\label{sec:streaming_student_details}
The full-sequence Teacher performs bidirectional multi-step denoising over an entire clip. For incremental generation, we distill it into a few-step block-causal Streaming Student. Following StreamAvatar~\citep{Sun_2026_CVPR}, the Student uses bidirectional attention within each four-latent window and causal attention across windows. We further adapt the dual-time flow maps and on-policy distribution-matching objective of AnyFlow~\citep{gu2026anyflow}.
For the $m$-th noisy window $\mathbf{x}^{(m)}_t$, destination time $r$, condition $\mathbf{c}^{(m)}$, and persistent state $\mathcal{S}_m$, the Student predicts
\begin{equation}
\begin{aligned}
\mathbf{v}_{\theta}\!\left(
\mathbf{x}^{(m)}_t,t,r;\mathbf{c}^{(m)},\mathcal{S}_m\right),\qquad
\mathbf{c}^{(m)}
=
\left(\mathbf{R}^{\mathrm{ref}},
\mathbf{A}^{(m)},
\mathbf{L}^{\mathrm{vtp}}\right).
\end{aligned}
\end{equation}
Training proceeds in two phases: the first learns source-destination flow maps, while the second performs on-policy chunked rollouts with distribution matching and the flow-map objective.

Our Prefix Streaming mechanism keeps each denoising window at four latent positions. The first window contains the clean reference latent and three new latents. Each later window uses the preceding chunk's final clean latent as a fixed prefix before generating three new latents:
\begin{equation}
\begin{aligned}
\mathbf{X}^{(0)}
&=[\mathbf{R}^{\mathrm{ref}},\mathbf{Z}^{(0)}_{1:3}],\\
\mathbf{X}^{(m)}
&=[\operatorname{sg}(\widehat{\mathbf{Z}}^{(m-1)}_3),
\mathbf{Z}^{(m)}_{1:3}],\quad m>0.
\end{aligned}
\end{equation}
The acoustic condition follows the same overlap pattern. Noisy intermediate states are never written to the persistent cache. After each window is denoised, an additional forward pass at $t=r=0$ rebuilds the cache from clean latents, while the duplicated prefix position is removed. Thus, each window advances the stream by exactly three latent positions.

Once the VTP is generated, speech and video are streamed using the same acoustic units. Every six units at 12.5 Hz produce a 0.48-second waveform segment and condition 12 video frames at 25 FPS:
\begin{equation}
\mathbf{U}_{6m+1:6m+6}
\longrightarrow
V_{12m+1:12m+12}.
\end{equation}
The reference and overlap latents are internal conditions and are not emitted as repeated frames. A final incomplete block is padded for computation and cropped to its valid duration. 

\paragraph{Prefix and Cache Implementation.}
\label{sec:prefix_cache_details}
For each post-initial window, noise is sampled for all four latent positions before the first position is replaced by the preceding clean latent. This prefix, together with its aligned acoustic condition, is re-injected after every denoising transition. After the final transition, a clean-cache forward
at $t=r=0$ updates the persistent state:
\begin{equation}
\begin{aligned}
\widehat{\mathbf{x}}^{(m)}_0
&=\operatorname{FlowMap}_{\theta}\!\left(
\mathbf{x}^{(m)}_{t_0};
t_0\!\rightarrow\!\cdots\!\rightarrow\!0,\mathcal{S}_m\right),\\
\mathcal{S}_{m+1}
&=\operatorname{CLEANCache}_{\theta}\!\left(
\widehat{\mathbf{x}}^{(m)}_0,\mathcal{S}_m\right).
\end{aligned}
\end{equation}
The duplicated prefix position is removed before updating the cache, so only the three newly generated positions are appended. 

\paragraph{Rate-Aligned Streaming Chunks.}
\label{sec:chunk_alignment}
The speech interface produces multi-codebook units at 12.5 Hz, while video is decoded at 25 FPS; hence, each speech unit aligns with two video frames. We group six units into a 0.48-second block, corresponding to 12 output frames or three new VAE latent positions: $\frac{6}{12.5}=0.48~\text{s}$. The overlap latent is used only as an internal prefix and is excluded from the decoded output. The final partial block is padded for computation, after which both waveform and video outputs are cropped to the valid duration. The same chunk definition is used during Student post-training and inference.

\paragraph{Optimization Objectives.}
\label{sec:optimization_details}
We train the framework with different objectives according to the available supervision in each stage. Let $\mathbf{o}^{\ast}=(o^{\ast}_1,\ldots,o^{\ast}_K)$ denote the complete structured assistant sequence containing the target VTP $\mathbf{p}^{\ast}$ and user-facing response $\mathbf{y}^{\ast}$. The language-modeling loss supervises both spans autoregressively:
\begin{equation}
    \mathcal{L}_{\mathrm{lm}}
    = -\sum_{k=1}^{K}
    \log p\left(o_k^{\ast} \mid o_{<k}^{\ast},
    \mathbf{E}_{\mathrm{in}}\right).
\end{equation}
The speech generator is conditioned only on hidden states and text tokens from the response span. Let $\mathbf{U}^{\ast}\in\mathbb{N}^{N\times C}$ denote the target multi-codebook sequence. Its loss includes the autoregressive first codebook and the residual codebooks predicted within each acoustic frame:
\begin{equation}
    \mathcal{L}_{\mathrm{sp}}
    = -\sum_{n=1}^{N}\sum_{c=1}^{C}
    \log p\!\left(
    u_{n,c}^{\ast}\,\middle|\,
    \begin{gathered}
      \mathbf{U}_{<n}^{\ast},\mathbf{u}_{n,<c}^{\ast}, \\
      \mathbf{H}^{\ell}_{\mathbf{y}},\mathbf{y}^{\ast},
      \mathbf{s}^{\mathrm{ref}}
    \end{gathered}
    \right).
\end{equation}
The VTP requires no separate regression loss because it is directly supervised as part of $\mathcal{L}_{\mathrm{lm}}$. During video training, the target VTP is serialized and encoded by the frozen video text encoder to obtain $\mathbf{L}^{\mathrm{vtp}}$. Following the flow-matching parameterization, the denoising model predicts the flow target $\boldsymbol{\epsilon} - \mathbf{x}_0$ from the noised latent and the three video conditions:
\begin{equation}
    \mathcal{L}_{\mathrm{vid}} =
    \left\|D_{\theta}(\mathbf{x}_{\sigma}, \sigma;
    \mathbf{R}^{\mathrm{ref}}, \mathbf{A},
    \mathbf{L}^{\mathrm{vtp}})
    - (\boldsymbol{\epsilon} - \mathbf{x}_0)\right\|_2^2.
\end{equation}
Across the heterogeneous training stages, the framework-level objective can be summarized as
\begin{equation}
    \mathcal{L} =
    \mathcal{L}_{\mathrm{lm}}
    + \mathcal{L}_{\mathrm{sp}}
    + \mathcal{L}_{\mathrm{vid}},
\end{equation}
where each term is enabled for the pathway and supervision available in the current batch, allowing the stages to use speech, dialogue, and video data without requiring all three annotations in every example.

\section{Data Construction}
\label{sec:appendix_data_construction}
\paragraph{ASR and TTS Data.}
Our Stage-1 speech data contain about 800K ASR examples and 1M TTS examples. The ASR portion consists of roughly 300K examples from LibriSpeech \citep{7178964} and 500K randomly sampled examples from Emilia \citep{10832365}. For TTS supervision, we randomly sample 1M text examples from Emilia and synthesize speech with Qwen3-TTS \citep{hu2026qwen3ttstechnicalreport}, matching the tokenizer and generator family used by the speech pathway.

\paragraph{Dialogue Data.}
For Stage-2 omni-modal response adaptation, we use examples from InstructS2S-200K~\citep{DBLP:conf/iclr/FangGZMZ025} and OmniCharacter~\citep{zhang-etal-2025-omnicharacter}. InstructS2S-200K provides speech-to-speech conversational supervision, while OmniCharacter contributes character-grounded, multi-turn dialogue examples. We construct target VTPs offline with Qwen3-VL-235B-A22B-Instruct: the model receives the reference image together with the dialogue context and target response, then fills the same five VTP fields used by the video pathway. These annotations supervise the structured \texttt{<thinking>}--\texttt{<response>} protocol without using target response videos. Together, the dialogue examples update the LLM, Speech Projector, and Speech Generator, adapting the shared speech interface toward interactive and character-consistent responses.

\paragraph{Video Generation Data.}
For the avatar-realization pathway in Stage~3, we use SpeakerVid \citep{zhang2025speakervid} as the source video dataset. We apply the automated filtering pipeline described below, obtaining about 140K training clips. For every retained clip, Qwen3-VL-235B-A22B-Instruct \citep{bai2025qwen3} constructs a five-field VTP describing the first frame, scene, emotion, movement style, and motion details using the prompt in Appendix~\ref{sec:prompt_templates}. We then tokenize its aligned speech with the Qwen3-TTS Tokenizer \citep{hu2026qwen3ttstechnicalreport} to obtain the multi-codebook video condition. Each retained clip becomes an avatar-realization record for training the diffusion model to map a reference image, VTP annotation, and aligned speech units to target video.

\begin{table*}[htbp]
  \centering
  \caption{Filtering pipeline for avatar-video training data.}
  \label{tab:video_data_filtering}
  \scriptsize
  \setlength{\tabcolsep}{3.0pt}
  \renewcommand{\arraystretch}{1.10}
  \begin{tabularx}{\textwidth}{
    >{\hsize=0.45\hsize\centering\arraybackslash}X
    >{\hsize=0.80\hsize\raggedright\arraybackslash}X
    >{\hsize=1.20\hsize\raggedright\arraybackslash}X
    >{\hsize=1.55\hsize\raggedright\arraybackslash}X}
    \toprule
    \textbf{Step} & \textbf{Filter} & \textbf{Model or Algorithm} & \textbf{Main Acceptance Rule} \\
    \midrule
    2a & Scene continuity & PySceneDetect \citep{castellano_pyscenedetect} AdaptiveDetector & No scene cut \\
    2b & Face visibility & YOLOv8x-face \citep{lindevs_yolov8_face} + IoU tracking & One visible face with $\geq98\%$ track coverage \\
    2c & Pose and motion & YOLOX-L \citep{yolox2021} + DWPose \citep{yang2023effective} & Complete facial landmarks and stable non-static motion \\
    2d & Audio--visual sync & S3FD \citep{zhang2017s3fdsingleshotscaleinvariant} + SyncNet \citep{Chung16a, 10.1145/3394171.3413532} & Sync-C $\geq7.0$ and Sync-D $\leq10.0$ \\
    2e & Visual quality & DOVER \citep{wu2023dover} + codec metadata & Height $\geq480$ and fused DOVER $\geq0.5$ \\
    2f & Exposure and flicker & OpenCV \citep{opencv} & No prolonged under/overexposure or strong flicker \\
    2g & Sharpness & HyperIQA \citep{Su_2020_CVPR} & HyperIQA $\geq55$ \\
    \bottomrule
  \end{tabularx}
\end{table*}

\paragraph{SpeakerVid Filtering and Record Construction.}
\label{sec:video_data_pipeline}
Audio segmentation first produces an aligned 16-kHz waveform and a 25-FPS video clip. We then apply the seven filters in Table~\ref{tab:video_data_filtering}, retaining clips that pass every check. For each retained clip, the first frame serves as the appearance reference; Qwen3-VL-235B-A22B-Instruct \citep{bai2025qwen3} annotates the five-field VTP, and the Qwen3-TTS Tokenizer \citep{hu2026qwen3ttstechnicalreport} converts the aligned speech into 16 codebook streams at 12.5 Hz. The final diffusion record contains the target video, reference frame, VTP, and speech tokens. At 25 FPS, each speech-token frame aligns with exactly two video frames, supplying the conditions required by the avatar generator.

\section{Implementation Details}
\label{sec:training_details}
\paragraph{Framework Training Strategy.}
\label{sec:appendix_training_strategy}
We use three data-oriented stages followed by one deployment-oriented distillation stage.
\paragraph{Stage 1: Speech Interface Alignment.}
We first use approximately 800K ASR and 1M TTS examples to establish the bidirectional speech interface while freezing the LLM, Speech Encoder, and Vision Encoder. ASR batches update the Speech Projector, grounding speech inputs in the LLM representation space. TTS batches update the Qwen3-TTS-initialized \citep{hu2026qwen3ttstechnicalreport} Speech Generator, teaching it to predict native multi-codebook units from response-side representations. Thus, Stage~1 updates the Speech Projector and Speech Generator, with each batch activating only the branch for which it provides supervision.

\paragraph{Stage 2: Omni-modal Response Adaptation.}
We then jointly update the LLM, Speech Projector, and Speech Generator using examples from InstructS2S-200K~\citep{DBLP:conf/iclr/FangGZMZ025} and OmniCharacter~\citep{zhang-etal-2025-omnicharacter}. Qwen3-VL-235B-A22B-Instruct provides the target VTP for each dialogue example from the reference image, dialogue context, and target response. The LLM learns the structured \texttt{<thinking>}--\texttt{<response>} protocol, where the thinking block contains the VTP, while speech supervision remains restricted to the response span. This stage adapts the aligned speech interface toward interactive and character-consistent spoken replies without requiring a target response video.

\paragraph{Stage 3: Avatar Video Realization.}
Finally, we train the full-sequence avatar diffusion model on the 140K SpeakerVid-derived records described above. Each training example provides a reference image, an automatically annotated VTP, teacher-forced multi-codebook speech units, and a target video. The diffusion model is therefore optimized to realize the same semantic and acoustic-temporal interfaces that the response pathway produces at inference time. The interface-based decomposition lets the response and video pathways use their respective data sources: dialogue supervision teaches what to say and how to plan a response, whereas SpeakerVid supervision teaches how to render a supplied plan and speech-unit sequence as video.

\paragraph{Stage 4: Streaming Student Distillation.}
After Stage~3, the full-sequence Video Generator is frozen as the Teacher. Stage~4 trains the Streaming Student in two successive phases. Phase~I learns source--destination flow maps from the Teacher, and Phase~II performs on-policy chunked Student rollouts with distribution matching while retaining the flow-map objective. Table~\ref{tab:framework_training_config} summarizes the optimization settings for all stages. For the video pathway, Stage~3 uses 125-frame SpeakerVid clips assigned to the $400\times720$, $720\times400$, and $720\times720$ buckets, while both Stage~4 phases use 121-frame windows with the same buckets.

\paragraph{Hyperparameters.}
We implement and train the framework with PyTorch 2.7.0 and CUDA 12.6. Each GPU has 16,896 CUDA cores, 132 streaming multiprocessors, and 141 GB of memory. Table~\ref{tab:framework_training_config} summarizes the stage-wise optimization settings and the two-phase Student distillation protocol; the corresponding data and trainable modules are specified in the stage descriptions above. Table~\ref{tab:inference_config} reports the inference configuration used for the main quality evaluation. Both models use empty negative prompts and deterministic dialogue-response decoding (temperature 0 and top-$p$ 1), while all Student results use the same validation-selected Prefix-Streaming checkpoint and differ only in their denoising budgets.

Video clips are decoded at 25 FPS. We assign each clip to its nearest aspect-ratio bucket among $400\times720$, $720\times400$, and $720\times720$, resize isotropically, and center-crop to the exact bucket without padding. A random contiguous window of 125 frames is used for Teacher SFT and 121 frames for Streaming Student distillation; short clips repeat their final frame. Window starts are aligned to the 12.5-Hz speech-token grid. For CommonEval, each sample uses a fixed reference image taken from the first frame and its speaker-matched reference speech. Reference speech is converted to mono, resampled to 24 kHz for the Speech Generator, and deterministically cropped to 3--10 seconds using seed 42; query speech and video-conditioning audio are mono at 16 kHz.

\begin{table*}[ht]
  \centering
  \caption{Unified optimization configuration. The upper panel compares the stage-wise device, batch, budget, learning-rate, and schedule settings; the lower panel reports the detailed hyperparameters of the two Stage~4 distillation phases.}
  \label{tab:framework_training_config}
  \scriptsize
  \setlength{\tabcolsep}{4.5pt}
  \renewcommand{\arraystretch}{1.14}
  \begin{tabularx}{\textwidth}{lccccc}
    \toprule
    \textbf{Setting} & \textbf{Stage 1} & \textbf{Stage 2} & \textbf{Stage 3} & \textbf{Stage 4, Phase I} & \textbf{Stage 4, Phase II} \\
    \midrule
    GPUs & 8 & 8 & 24 & 40 & 40 \\
    Batch & 16 & 1 & 1 & 1 & 1 \\
    Budget & 1 epoch & 2 epochs & 10K steps & 6K steps & 3K steps \\
    LR & Proj.: $1\times10^{-3}$; Gen.: $1\times10^{-4}$ & LLM: $1\times10^{-6}$; Proj./Gen.: $2\times10^{-5}$ & $2\times10^{-5}$ & $5\times10^{-5}$ & $2\times10^{-6}$ \\
    Schedule & Cosine; 30\% warmup & Cosine; 10\% warmup & Constant & 1K-step warmup & No warmup \\
    \bottomrule
  \end{tabularx}

  \vspace{5pt}
  \textbf{Stage 4 distillation details.} The stream state is reported as new/overlap/sink/rolling latent frames.
  \vspace{2pt}
  \setlength{\tabcolsep}{30pt}
  \begin{tabularx}{\textwidth}{
    >{\hsize=1.40\hsize\raggedright\arraybackslash}X
    >{\hsize=0.80\hsize\centering\arraybackslash}X
    >{\hsize=0.80\hsize\centering\arraybackslash}X}
    \toprule
    \textbf{Hyperparameter} & \textbf{Phase I} & \textbf{Phase II} \\
    \midrule
    Diffusion horizon / sigma shift & 1,000 / 5.0 & 1,000 / 5.0 \\
    Preferred inference steps & 4 & 4 \\
    Rollout step set & -- & $\{2,4,8,16\}$ \\
    Gate / destination-time type & 0.25 / $r$ & 0.25 / $r$ \\
    Diffusion / consistency weights & 0.5 / 0.25 & 0.5 / 0.25 \\
    Flow-map guidance / $\epsilon$ & 3.0 / 5.0 & 3.0 / 5.0 \\
    DMD / co-training weights & -- & 0.5 / 1.0 \\
    Real-score guidance & -- & 2.5 (CFG 3.5) \\
    Text drop / bidirectional prob. & 0.1 / 0.1 & 0.1 / 0.0 \\
    Supervised chunks / long-context ratio & 3 / 0.5 & 3 / 1.0 \\
    Stream state / max relative-position distance & 3/1/4/6 / 9 & 3/1/4/6 / 9 \\
    \bottomrule
  \end{tabularx}
\end{table*}

\begin{table}[ht]
  \centering
  \caption{Video inference protocol used for the main quality experiments. Text/audio CFG lists the two guidance scales in that order. Student output length follows the generated speech duration.}
  \label{tab:inference_config}
  \scriptsize
  \setlength{\tabcolsep}{8pt}
  \renewcommand{\arraystretch}{1.09}
  \begin{tabularx}{0.54\columnwidth}{lcc}
    \toprule
    \textbf{Setting} & \textbf{Teacher} & \textbf{Student} \\
    \midrule
    Temporal attention & Bidirectional & Block causal \\
    Output frames & 125 & Speech aligned \\
    Denoising steps & 50 & 2, 4, or 8 \\
    Scheduler & FlowMatch & FlowMap discrete \\
    Training horizon / sigma shift & 1,000 / 5.0 & 1,000 / 5.0 \\
    Text/audio CFG & 3.5 / 8.5 & 1.0 / 1.0 \\
    Generation/prompt seed & 42 / 42 & 42 / 42 \\
    Frame rate & 25 FPS & 25 FPS \\
    \bottomrule
  \end{tabularx}
\end{table}

\section{Prompt Templates}
\label{sec:prompt_templates}

\paragraph{VTP Annotation Prompt.}
The following prompt is provided to Qwen3-VL-235B-A22B-Instruct \citep{bai2025qwen3} together with each retained training video.

\begin{Verbatim}[fontsize=\scriptsize,breaklines=true,breakanywhere=true,breaksymbolleft={},breaksymbolright={}]
You are a meticulous video annotator for speech-driven human avatar generation data.

Watch the video carefully and describe visible appearance, speaking behavior, emotional expression, and human motion in a generation-friendly manner.

Global annotation rules:
1. Describe only clearly visible and visually verifiable content.
2. Use concise, concrete, generation-oriented language.
3. Prefer observable motion and appearance over abstract interpretation.

Avoid: inferred intentions or beliefs, invisible actions, story-like narration, cinematic language, or unsupported semantic speculation.

Note:
1. Output ONLY XML-style tags in the exact order below.
2. Each tag must appear exactly once.
3. Do not add explanations or extra text outside the XML tags.

Template:
<first_frame_scene>
One sentence describing only the first frame: appearance, framing (e.g., close-up, medium shot, full body), pose, facial expression, background, and lighting.
</first_frame_scene>

<scene>
One sentence describing the overall situation, atmosphere, and visible background activity when present.
</scene>

<emotion>
One or two emotion labels chosen from: neutral, happy, sad, angry, surprised, fearful, disgusted, calm, concerned, confident, excited, nervous, thoughtful, engaged, amused.
Compound emotions are allowed using comma separation. If the emotion shifts during the video, use an arrow (e.g., neutral -> excited).
</emotion>

<movement_style>
A short phrase describing the high-level motion style, overall energy level, expressiveness, and vibe (e.g., "formal and restrained", "energetic and theatrical"). Do not describe specific body parts here.
</movement_style>

<motion_description>
One or two sentences detailing the specific physical kinematics: visible head, facial, hand, shoulder, torso, and body movements, including their directions and approximate amplitudes.
</motion_description>
\end{Verbatim}

\paragraph{Dialogue-data VTP Construction.}
For Stage~2 dialogue data, Qwen3-VL-235B-A22B-Instruct \citep{bai2025qwen3} receives the reference image, the available role or dialogue context, and the target response. The instruction asks it to fill the same five XML fields as above, but to describe a plausible response-specific visual plan rather than an observed target video. The first-frame field describes the supplied reference image, while the scene, emotion, movement style, and motion-description fields describe the intended speaking behavior for the target response. No target response video is provided for these dialogue examples.

\paragraph{VTP Visual-Grounding Judge Prompt.}
For the main protocol in Table~\ref{tab:vtp_visual_alignment}, Gemini~3.1~Pro receives the following system prompt. The first supplied image is the reference image; the remaining inputs are chronological generated-video samples excluding frame~0.

\begin{Verbatim}[fontsize=\scriptsize,breaklines=true,breakanywhere=true,breaksymbolleft={},breaksymbolright={}]
You are a strict reference-image and video-to-plan alignment evaluator.
Judge only visible evidence in the supplied images. Do not assume that a plan is correct merely because it is plausible. The first supplied image is the ORIGINAL REFERENCE IMAGE. Use it only to judge first_frame_scene. All remaining images are chronological GENERATED-video samples; video frame 0, which repeats the reference image, has been excluded. Use only these generated-video samples to judge scene, emotion, movement_style, and motion_description.

For each of the five VTP fields, assign exactly one binary point:
- 1: the field's central visual meaning is visibly supported.
- 0: the central meaning is contradicted, absent, or too weak to verify.

Judge at field level, not by exact word matching. Do not fail an otherwise aligned field because of a minor omitted prop, crop, paraphrase, or left/right ambiguity caused by camera viewpoint or mirroring. For a field with several details, award the point when its principal subject, setting, affect, style, or action is preserved and there is no major semantic contradiction.

For movement fields, require visible temporal evidence across generated-video samples. For first_frame_scene, judge only the original reference image and focus on its principal subject, setting, and pose. Return JSON only.
\end{Verbatim}

\section{Evaluation Protocol and Metric Details}
\label{sec:eval_metric_details}
\paragraph{Baselines and Comparison Protocol.}
Table~\ref{tab:main_results} groups the audio-video comparison by supported interface: video-only renderers (echomimic \citep{chen2024echomimic}, StableAvatar-1.3B \citep{DBLP:journals/corr/abs-2508-08248}, OmniAvatar-1.3B \citep{DBLP:journals/corr/abs-2506-18866}, and FantasyTalking \citep{wang2025fantasytalking}), video-audio generators (Universe-1 \citep{Universe_1} and UniAVGen \citep{zhang2026uniavgenunifiedaudiovideo}), and Ex-Omni-2D Teacher and Streaming Student variants. Table~\ref{tab:omnicharacter_comparison} compares against representative role-playing and dialogue models from the OmniCharacter protocol, including ChatGLM3-6B \citep{glm2024chatglm}, Baichuan2-7B \citep{baichuan2023baichuan2}, InternLM-7B \citep{cai2024internlm2}, CharacterGLM-6B \citep{zhou2023characterglm}, Llama-3.1-8B \citep{grattafiori2024llama}, OmniCharacter \citep{zhang-etal-2025-omnicharacter} and Qwen-family baselines \citep{DBLP:journals/corr/abs-2309-16609, DBLP:journals/corr/abs-2412-15115, yang2025qwen3, DBLP:journals/corr/abs-2407-10671, DBLP:journals/corr/abs-2503-20215}. Table~\ref{tab:speech_qa} compares with speech-enabled omni-modal systems on VoiceBench \citep{chen2024voicebenchbenchmarkingllmbasedvoice}: Qwen2.5-Omni \citep{DBLP:journals/corr/abs-2503-20215}, Moshi \citep{DBLP:journals/corr/abs-2410-00037}, VITA-1.5 \citep{fu2026vita}, Mini-Omni2 \citep{DBLP:journals/corr/abs-2410-11190}, and SLAM-Omni \citep{chen2025slam}.

For audio-video baselines that require an upstream response, Qwen2.5-Omni-7B is used as the common response controller, whose parameter scale is close to our 8B LLM backbone. A2V renderers receive speech from this controller, so PQ and CU are not attributed to the downstream renderer; SIM is reported only when the evaluated interface supports reference speech. UniAVGen's main video metrics use A2V mode because its joint A-V mode supports only 5-second outputs, while its SIM score is obtained separately in RAVG reference-speech mode. All Ex-Omni-2D rows share the same generated speech and therefore have identical audio metrics. Teacher and Streaming Student quality comparisons reuse the same text response, VTP, speech units, waveform, reference image, and reference speech, so their reported differences isolate the video generator. Table~\ref{tab:main_results} evaluates practical capability under each supported interface, and Appendix Table~\ref{tab:upstream_cascade_ablation} diagnoses the effect of changing the upstream response source while keeping each downstream renderer fixed.

\paragraph{Audio Metrics.}
We use AudioBox-Aesthetics \citep{tjandra2025aes} to evaluate generated speech with PQ and CU. PQ focuses on perceptual production quality, reflecting whether the audio sounds clear, natural, stable, and free of obvious artifacts. CU evaluates the practical usefulness of the generated content, emphasizing intelligibility and semantic completeness. Compared with strict word-error-rate matching, CU better captures whether the generated speech is usable and understandable from a listener's perspective. We additionally report SIM using the official Seed-TTS \citep{anastassiou2024seed} evaluation protocol: WavLM-Large \citep{Chen2021WavLM} and an ECAPA-TDNN \citep{desplanques2020ecapa} speaker-verification head embed the generated and reference speech, and SIM is their cosine similarity.

\paragraph{Video Metrics.}
We use VBench \citep{huang2023vbench} to evaluate generated videos with SC, IQ, and DD; none of these metrics requires a ground-truth video. SC measures the temporal consistency of the generated subject, IQ measures perceptual frame quality such as sharpness and visible artifacts, and DD uses optical-flow magnitude to report the proportion of videos with substantial motion. A higher DD indicates more frequent motion, but does not necessarily imply better motion quality or semantic appropriateness.

\paragraph{A-V Sync.}
We use SyncNet confidence (Sync-C) \citep{Chung16a, 10.1145/3394171.3413532} to evaluate whether mouth movements are temporally aligned with the generated speech. Sync-C measures the separation between the best audio-video alignment and competing temporal offsets, so higher values indicate more confident synchronization.

\paragraph{Speech QA Metrics.}
For spoken question answering, we use AlpacaEval \citep{alpaca_eval}, CommonEval \citep{chen2024voicebenchbenchmarkingllmbasedvoice}, and BBH \citep{suzgun2022challenging} from VoiceBench \citep{chen2024voicebenchbenchmarkingllmbasedvoice} and follow the open-source VoiceBench evaluation code. AlpacaEval and CommonEval use GPT \citep{openai2024gpt4omini} to judge open-ended response quality on a five-point scale, where 5 is the best score; AlpacaEval emphasizes general instruction following, whereas CommonEval focuses on common spoken QA. BBH evaluates complex QA requiring multi-step reasoning, logical composition, and abstract thinking, and is reported as accuracy. Each BBH subtask then applies task-specific string matching or regular expressions to convert free-form responses into binary labels, such as yes/no answers, option choices, navigation decisions, truthfulness judgments, or sports-plausibility judgments. The final score is computed as the percentage of correct predictions.

\section{Additional Analysis}
\label{sec:additional_dialogue_results}
\paragraph{Backbone Capability Comparison.}
To analyze how multimodal adaptation affects the language capability inherited from initialization, we compare Ex-Omni-2D with its Qwen3-8B backbone under the same open-ended VoiceBench protocol. As shown in Table~\ref{tab:backbone_capability}, Ex-Omni-2D scores 4.48 on AlpacaEval and 4.35 on CommonEval with text input, compared with 4.82 and 4.51 for Qwen3-8B. The relatively small reductions indicate that Ex-Omni-2D largely retains the backbone's text capability, while incurring a modest trade-off after multimodal adaptation. This degradation may result from interference when the shared language backbone is optimized across heterogeneous modalities and tasks. 

\begin{table}[htbp]
  \centering
  \caption{Backbone capability comparison}
  \label{tab:backbone_capability}
  \scriptsize
  \setlength{\tabcolsep}{13.5pt}
  \begin{tabularx}{0.54\columnwidth}{lcc}
    \toprule
    \textbf{Method} & \textbf{AlpacaEval} $\uparrow$ & \textbf{CommonEval} $\uparrow$ \\
    \midrule
    Qwen3-8B (text input) & \textbf{4.82} & \textbf{4.51} \\
    Qwen3-8B (speech input) & N/A & N/A \\
    \midrule
    Ours (text input) & 4.48 & 4.35 \\
    Ours (speech input) & 4.28 & 3.71 \\
    \bottomrule
  \end{tabularx}
\end{table}

\paragraph{Multi-turn Context Ablation.}
\label{sec:omnicharacter_context_ablation}
We isolate dialogue history on the 400-dialogue OmniCharacter test set. Both settings receive the benchmark Role Card and are scored on the final turn with the same CharacterEval reward model. The context-free setting receives only the final user query, whereas the full setting additionally receives all preceding question--answer turns. Table~\ref{tab:omnicharacter_context_ablation} shows higher scores in every evaluated dimension when dialogue history is provided, with especially large differences in persona behavior, persona utterance, expression diversity, and response consistency. Across the twelve dimensions, the dialogue-level macro average increases from 2.769 (95\% CI: 2.725--2.812) to 3.283 (3.242--3.322); the paired difference is 0.514 with a 95\% bootstrap CI of 0.464--0.562. Under this evaluation protocol, preceding dialogue history is therefore associated with a consistent improvement over using only the final query.

\begin{table*}[htbp]
  \centering
  \caption{Multi-turn context ablation on OmniCharacter. KE: Knowledge-Exposure, KA: Knowledge-Accuracy, KH: Knowledge-Hallucination, PB: Persona-Behavior, PU: Persona Utterance, Flu.: Fluency, Coh.: Coherency, Cons.: Consistency, HL: Human-Likeness, CS: Communication Skill, ED: Expression Diversity, and Emp.: Empathy. Both settings retain the same Role Card and differ only in whether preceding dialogue turns are provided.}
  \label{tab:omnicharacter_context_ablation}
  \scriptsize
  \setlength{\tabcolsep}{5.2pt}
  \renewcommand{\arraystretch}{1.10}
  \begin{tabularx}{\textwidth}{lccccccccccccc}
    \toprule
    \textbf{Model} & \textbf{KE} $\uparrow$ & \textbf{KA} $\uparrow$ & \textbf{KH} $\uparrow$ & \textbf{PB} $\uparrow$ & \textbf{PU} $\uparrow$ & \textbf{Flu.} $\uparrow$ & \textbf{Coh.} $\uparrow$ & \textbf{Cons.} $\uparrow$ & \textbf{HL} $\uparrow$ & \textbf{CS} $\uparrow$ & \textbf{ED} $\uparrow$ & \textbf{Emp.} $\uparrow$ & \textbf{Avg.} $\uparrow$ \\
    \midrule
    w/o dialogue context & 1.798 & 2.864 & 2.461 & 2.035 & 2.583 & 3.461 & 3.743 & 3.410 & 3.198 & 2.987 & 1.535 & 3.145 & 2.769 \\
    w/ dialogue context & \textbf{2.076} & \textbf{3.410} & \textbf{2.994} & \textbf{2.911} & \textbf{3.354} & \textbf{3.812} & \textbf{4.100} & \textbf{3.902} & \textbf{3.893} & \textbf{3.362} & \textbf{2.239} & \textbf{3.341} & \textbf{3.282}\\
    \bottomrule
  \end{tabularx}
\end{table*}

\paragraph{Streaming Student Denoising-Step Analysis.}
\label{sec:student_denoising_steps}

We compare denoising-step counts using the same validation-selected Prefix-Streaming checkpoint, 200 CommonEval requests, reference images, VTPs, speech units, waveforms, and random seed as Table~\ref{tab:main_results}; only the inference-time denoising budget changes. The speech pathway is invariant, so we report only video and synchronization metrics. Increasing the denoising budget consistently increases SC, IQ, DD, and Sync-C in Table~\ref{tab:student_denoising_sweep}; for DD, this means a larger fraction of outputs with substantial motion. Two steps provide the fastest tested operating point and the lowest reported quality metrics; four steps provide an intermediate operating point; and eight steps obtain the highest SC, IQ, DD, and Sync-C among the Student settings at lower throughput. The Student--Teacher gap quantifies the quality cost of distilling a 50-step bidirectional Teacher into a compact 1.3B block-causal model with only a few denoising steps. The corresponding throughput and latency measurements are reported in Table~\ref{tab:streaming_latency}.

\begin{table}[htbp]
  \centering
  \caption{CommonEval quality when varying the number of Streaming Student denoising steps.}
  \label{tab:student_denoising_sweep}
  \scriptsize
  \setlength{\tabcolsep}{9.5pt}
  \renewcommand{\arraystretch}{1.08}
  \begin{tabularx}{0.48\columnwidth}{
    >{\hsize=1.4\hsize\centering\arraybackslash}c
    *{4}{>{\hsize=0.9\hsize\centering\arraybackslash}c}}
    \toprule
    \textbf{Denoising Steps} & \textbf{SC} $\uparrow$ & \textbf{IQ} $\uparrow$ & \textbf{DD} & \textbf{Sync-C} $\uparrow$ \\
    \midrule
    2 & 93.33 & 52.70 & 9.50 & 3.514 \\
    4 & 93.65 & 57.40 & 32.00 & 3.899 \\
    8 & 93.91 & 61.15 & 48.00 & 3.999 \\
    \bottomrule
  \end{tabularx}
\end{table}

\paragraph{Prefix-Streaming Ablation.}
\label{sec:prefix_comparison}

\begin{figure}[htbp]
  \centering
  \includegraphics[width=0.48\textwidth]{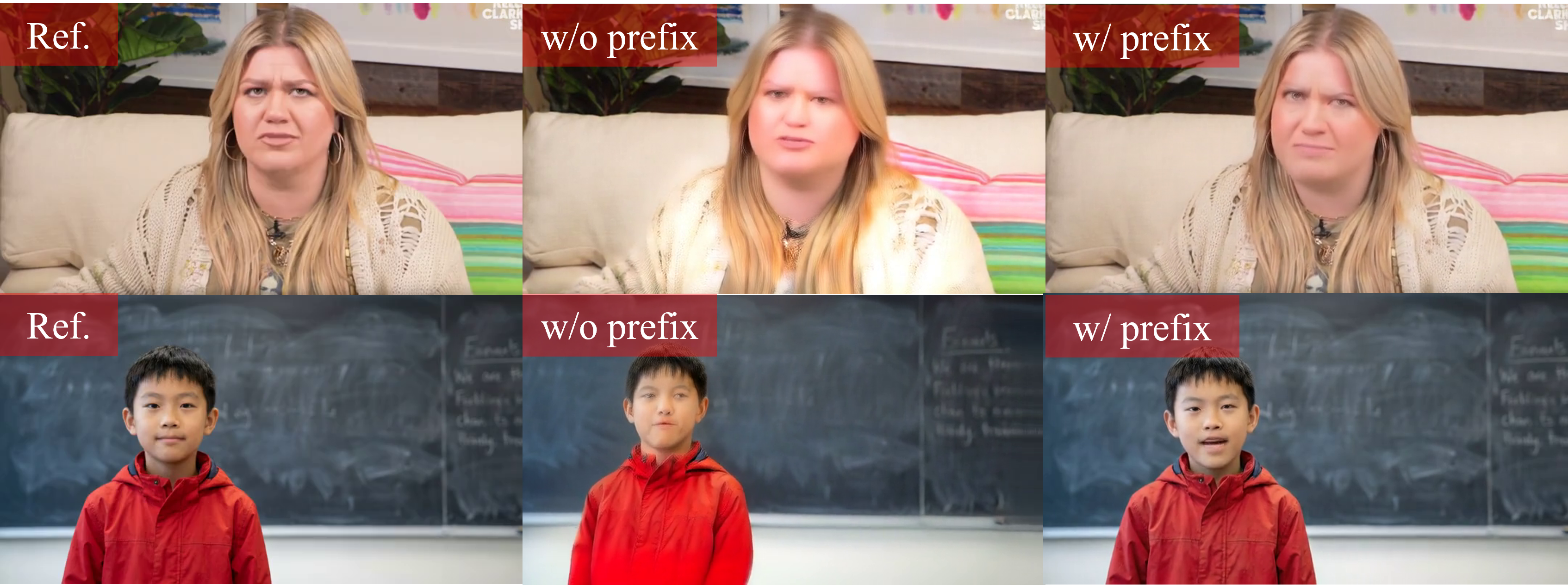}
  \vspace{2pt}
  \includegraphics[width=0.48\textwidth]{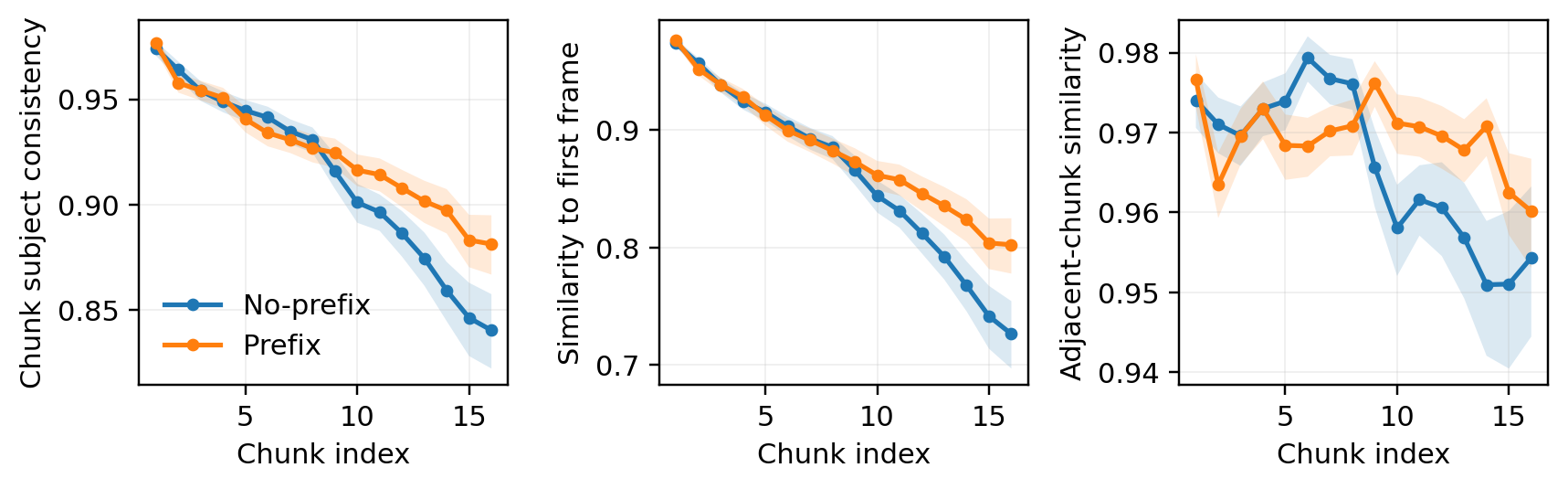}
  \caption{Prefix-Streaming analysis. \textbf{Top:} representative later-chunk examples; each row shows the reference image, a no-prefix frame, and its Prefix-Streaming counterpart. \textbf{Bottom:} full-duration DINO curves over 200 paired CommonEval videos. Markers denote chunk means and shaded regions denote paired bootstrap 95\% confidence intervals over videos. Prefix is mixed over early chunks but degrades more slowly and is consistently better from chunk~9 onward, consistent with reduced cumulative late-chunk subject drift rather than a uniform per-chunk improvement.}
  \label{fig:prefix_comparison}
\end{figure}

Figure~\ref{fig:prefix_comparison} compares our earlier no-prefix Streaming Student with the Prefix-Streaming Student using representative later-chunk examples and full-set temporal curves. In the examples, the no-prefix system shows visible facial or body degradation in later chunks, whereas the Prefix-Streaming system better preserves appearance. This comparison complements the cache analysis in Appendix~\ref{sec:streaming_student_details}: cached history alone does not place the preceding clean latent inside the current denoising window. As shown in the right figure in Figure~\ref{fig:prefix_comparison}, the 16-chunk DINO \cite{caron2021emerging} analysis increases mean chunk subject consistency from 0.9251 to 0.9319, reduces the consistency-error slope from 0.00783 to 0.00599 per chunk, and lowers last-minus-first consistency error from 0.1005 to 0.0790, a 21.4\% reduction. Prefix is mixed during the first eight chunks but is higher at every chunk from 9 through 16, with all paired 95\% confidence intervals excluding zero. This phenomenon shows that prefix can partially mitigates drift problem in video generation.

\paragraph{Long-Horizon Prefix Analysis.}
\label{sec:prefix_long_horizon}

We analyze whether Prefix Streaming improves stability as generation proceeds. The DINO curves in Figure~\ref{fig:prefix_comparison} measure overall subject consistency across all 16 chunks. We further track 106 facial landmarks \citep{insightface} after removing global translation, scale, and rotation. Facial acceleration and jerk are normalized by each video's first second, with lower ratios indicating less growth in irregular facial motion. As shown in Table~\ref{tab:prefix_face_dynamics}, Prefix Streaming consistently improves face-detection continuity and reduces acceleration and jerk ratios. After five seconds, face detection increases from 94.47\% to 97.23\%, while the acceleration and jerk ratios decrease from 2.277 to 1.884 and from 2.341 to 1.966, respectively. The paired 95\% bootstrap confidence intervals exclude zero for all three differences.

\begin{table*}[htbp]
  \centering
    \caption{Long-horizon facial stability on paired CommonEval videos. Facial acceleration and jerk are normalized by each video's first-second baseline; lower ratios indicate less growth in irregular facial motion. Confidence intervals are computed using 10,000 paired bootstrap samples.}
  \label{tab:prefix_face_dynamics}
  \scriptsize
  \setlength{\tabcolsep}{12.5pt}
  \renewcommand{\arraystretch}{1.10}
  \begin{tabularx}{\textwidth}{ccccc}
    \toprule
    \textbf{Tail at/after} & \textbf{Pairs} & \textbf{Face detection (No-prefix / Prefix)} & \textbf{Acceleration ratio (No-prefix / Prefix)} & \textbf{Jerk ratio (No-prefix / Prefix)} \\
    \midrule
    1 s & 200 & 97.44 / \textbf{98.73} & 1.923 / \textbf{1.730} & 1.949 / \textbf{1.783} \\
    2 s & 198 & 97.13 / \textbf{98.59} & 1.988 / \textbf{1.781} & 2.024 / \textbf{1.844} \\
    3 s & 182 & 96.50 / \textbf{98.32} & 2.077 / \textbf{1.772} & 2.124 / \textbf{1.845} \\
    4 s & 168 & 95.76 / \textbf{97.94} & 2.212 / \textbf{1.782} & 2.281 / \textbf{1.867} \\
    5 s & 145 & 94.47 / \textbf{97.23} & 2.277 / \textbf{1.884} & 2.341 / \textbf{1.966} \\
    \bottomrule
  \end{tabularx}
\end{table*}

\paragraph{User-Visible Streaming Latency.}
\label{sec:user_visible_latency}

Table~\ref{tab:user_visible_latency} reports event-level latency from request arrival for the four-GPU, four-step Prefix-Streaming pipeline. The measurements distinguish early internal planning from the first output a user can hear or play. Although VTP generation begins quickly, response planning and speech-unit generation delay the first audible speech to 2.308 seconds and the first playable 0.48-second video chunk to 3.142 seconds. The 0.834-second gap between first audio and first video is noticeable in an interactive interface and motivates tighter scheduling between speech-unit generation and video rendering. These measurements show that response planning and speech-unit generation dominate the startup time before video playback.

\begin{table}[htbp]
  \centering
  \caption{User-visible latency of the four-GPU, four-step Prefix-Streaming pipeline. Confidence intervals use bootstrap resampling over complete requests.}
  \label{tab:user_visible_latency}
  \scriptsize
  \setlength{\tabcolsep}{10.0pt}
  \renewcommand{\arraystretch}{1.08}
  \begin{tabularx}{0.48\columnwidth}{lcc}
    \toprule
    \textbf{Event from request arrival} & \textbf{Mean (s)} & \textbf{95\% CI (s)} \\
    \midrule
    First VTP token & 0.193 & [0.180, 0.207] \\
    VTP completion & 1.458 & [1.414, 1.504] \\
    First response token & 1.470 & [1.425, 1.515] \\
    First speech units & 2.250 & [2.188, 2.312] \\
    First PCM / audible speech & 2.308 & [2.247, 2.370] \\
    First playable video chunk & 3.142 & [3.067, 3.216] \\
    \bottomrule
  \end{tabularx}
\end{table}

\paragraph{Upstream Response Source Ablation.}
\label{sec:upstream_cascade_ablation}
Table~\ref{tab:upstream_cascade_ablation} extends the controlled upstream comparison to all downstream generators. Within each pair, we keep the requests, reference conditions, downstream checkpoint, and evaluation protocol fixed, and change only the upstream source. Consistent with the main-text analysis, the effect is mixed and renderer-dependent. EchoMimic and OmniAvatar change only slightly; StableAvatar and FantasyTalking obtain lower Sync-C, whereas Universe-1 and UniAVGen obtain higher Sync-C at the cost of decreases
in other metrics. FantasyTalking shows the largest SC drop, while UniAVGen shows the largest Sync-C increase. Nevertheless, OmniAvatar retains the highest Sync-C under both upstream sources. These results show that changing the upstream model affects individual scores but provides no consistent advantage across renderers. We therefore use Qwen2.5-Omni as the common upstream source for external generators in the main comparison and report these paired results as a sensitivity analysis.
\begin{table*}[htbp]
  \centering
  \caption{Ablation of the upstream response source for fixed downstream generators. A2V methods do not report upstream audio metrics. UniAVGen video metrics use its A2V mode because joint A-V generation is limited to 5-second outputs; its SIM is separately measured in RAVG reference-speech mode.}
  \label{tab:upstream_cascade_ablation}
  \scriptsize
  \setlength{\tabcolsep}{12.0pt}
  \renewcommand{\arraystretch}{1.08}
  \begin{tabularx}{\textwidth}{llccccccc}
    \toprule
    \textbf{Downstream Generator} & \textbf{Upstream Source} & \textbf{PQ} $\uparrow$ & \textbf{CU} $\uparrow$ & \textbf{SIM} $\uparrow$ & \textbf{SC} $\uparrow$ & \textbf{IQ} $\uparrow$ & \textbf{DD} & \textbf{Sync-C} $\uparrow$ \\
    \midrule
    \multirow{2}{*}{echomimic} & Qwen2.5-Omni & -- & -- & -- & 97.87 & 54.37 & 74.00 & 4.82 \\
      & Ex-Omni-2D & -- & -- & -- & 97.90 & 54.63 & 72.00 & 4.88 \\
    \midrule
    \multirow{2}{*}{StableAvatar-1.3B} & Qwen2.5-Omni & -- & -- & -- & 97.40 & 67.51 & 42.00 & 3.38 \\
      & Ex-Omni-2D & -- & -- & -- & 97.43 & 67.56 & 38.50 & 2.87 \\
    \midrule
    \multirow{2}{*}{OmniAvatar-1.3B} & Qwen2.5-Omni & -- & -- & -- & 97.74 & 66.65 & 15.50 & 5.64 \\
      & Ex-Omni-2D & -- & -- & -- & 98.19 & 66.89 & 22.50 & 5.48 \\
    \midrule
    \multirow{2}{*}{FantasyTalking} & Qwen2.5-Omni & -- & -- & -- & 96.78 & 64.21 & 6.50 & 3.43 \\
      & Ex-Omni-2D & -- & -- & -- & 93.19 & 63.60 & 8.00 & 3.20 \\
    \midrule
    \multirow{2}{*}{Universe-1} & Qwen2.5-Omni & 4.16 & 3.98 & -- & 97.35 & 67.72 & 34.00 & 1.75 \\
      & Ex-Omni-2D & 4.11 & 3.74 & -- & 97.21 & 67.60 & 29.00 & 1.94 \\
    \midrule
    \multirow{2}{*}{UniAVGen} & Qwen2.5-Omni & -- & -- & 0.291 & 96.07 & 68.02 & 55.00 & 4.15 \\
      & Ex-Omni-2D & -- & -- & 0.291 & 95.92 & 65.75 & 54.50 & 5.07 \\
    \bottomrule
  \end{tabularx}
\end{table*}

\paragraph{VTP Grounding and Observational Alignment.}
\label{sec:additional_vtp_analysis}
We additionally evaluate whether the generated VTP is visually grounded rather than merely structurally valid. On 200 CommonEval examples, a third-party multimodal judge assigns one binary score to each VTP field based on visible evidence. As shown in Table~\ref{tab:vtp_visual_alignment}, the \texttt{first\_frame\_scene} field obtains 43.00\%, lower than the other fields. We believe this in because that the field is evaluated against a single initial image, whereas the VTP describes an evolving visual response whose scene and motion may change over time.  The remaining fields are evaluated using nine chronological samples from the generated video and achieve an average alignment of 85.38\%, including 89.50\% for emotion and 94.00\% for movement style. Overall, the results show that Ex-Omni-2D generally translates its generated VTP into observable visual behavior. Judge prompts are provided in Appendix~\ref{sec:prompt_templates}.

\begin{table*}[htbp]
\centering
\caption{VTP grounding and observational alignment on 200 CommonEval
examples. First-frame is evaluated against the reference image; the remaining fields are evaluated on chronological samples from the generated video.}
\label{tab:vtp_visual_alignment}
\scriptsize
\setlength{\tabcolsep}{6pt}
\renewcommand{\arraystretch}{1.12}
\begin{tabular}{cccccc}
\toprule
\textbf{First-frame} & \textbf{Scene} & \textbf{Emotion}
& \textbf{Movement} & \textbf{Motion} & \textbf{Video Avg.} \\
\midrule
43.00 & 84.50 & 89.50 & 94.00 & 73.50 & 85.38 \\
\bottomrule
\end{tabular}
\end{table*}

\paragraph{VTP Motion-Field Intervention.}
To complement the observational alignment analysis above, we run a paired intervention that changes only the VTP motion fields. On 16 CommonEval examples, we fix the reference image, generated waveform, speech units, sampling configuration, random seed, and the first-frame, scene, and emotion VTP fields. We then generate two Teacher videos per example by replacing only \texttt{movement\_style} and \texttt{motion\_description}: the restrained condition asks for subtle lip movement, small natural head motion, and minimal upper-body or hand movement, while the expressive condition asks for larger facial-expression changes, frequent head motion, and visible upper-body or hand gestures when the framing allows. We score the paired outputs with two local motion diagnostics: mean adjacent-frame gray-scale difference and mean optical-flow magnitude \citep{DBLP:conf/eccv/TeedD20} over decoded frames. These metrics measure visible frame-to-frame motion and can also respond to artifacts or drift, so they are used as auxiliary diagnostics rather than semantic motion-quality scores. Table~\ref{tab:vtp_motion_intervention} shows that the expressive motion fields increase both diagnostics on average, and the expressive video has the larger value in 14 of 16 paired examples for both metrics. The paired bootstrap confidence intervals for the mean deltas are $[-0.0010,0.0057]$ for frame difference and $[-0.0754,0.3685]$ for optical-flow magnitude. This results shows that VTP motion fields can bias the Teacher toward stronger visible motion under fixed speech and reference conditions.

\begin{table}[htbp]
  \centering
  \caption{Paired VTP motion-field intervention on 16 short CommonEval examples. Only \texttt{movement\_style} and \texttt{motion\_description} are changed between conditions. ``Expressive higher'' reports the fraction of paired examples where the expressive condition has the larger metric value.}
  \label{tab:vtp_motion_intervention}
  \scriptsize
  \setlength{\tabcolsep}{8.0pt}
  \renewcommand{\arraystretch}{1.10}
  \begin{tabularx}{0.65\columnwidth}{lcccc}
    \toprule
    \textbf{Metric} & \textbf{Restrained} & \textbf{Expressive} & \textbf{$\Delta$} & \textbf{Expressive higher} \\
    \midrule
    Frame difference & 0.0135 & 0.0162 & +0.0026 & 14/16 \\
    Optical-flow magnitude & 0.7400 & 0.9071 & +0.1670 & 14/16 \\
    \bottomrule
  \end{tabularx}
\end{table}

\paragraph{Ablation on Speech Conditioning Pathway.}
\label{sec:speech_unit_interface}
As shown in Table~\ref{tab:speech_unit_interface}, we compare waveform--wav2vec conditioning \citep{DBLP:conf/nips/BaevskiZMA20} with native single- and 16-codebook interfaces on the same 200 CommonEval utterances. Waveform--wav2vec conditioning obtains Sync-C 5.83 but requires waveform decoding, resampling, and re-encoding. Single-codebook units reduce conditioning latency to 0.003 seconds but obtain Sync-C 2.07. The 16-codebook interface obtains Sync-C 4.95 with 0.011-second conditioning latency, making the measured trade-off explicit without requiring waveform reconstruction. The waveform baseline remains stronger for synchronization, but it conditions the video path on an already decoded and re-encoded audio signal; the unit interface is the lower-latency representation naturally emitted by the response model and is therefore used in the streaming pipeline.

\begin{table}[htbp]
  \centering
  \caption{Synchronization and conditioning latency of three video-conditioning interfaces. Latency starts from available speech units and ends before the first video denoising step. The waveform interface includes code2wav, in-memory resampling, and wav2vec encoding. Results are averaged over 200 CommonEval utterances on one GPU with the same hardware configuration.}
  \label{tab:speech_unit_interface}
  \scriptsize
  \setlength{\tabcolsep}{12.8pt}
  \renewcommand{\arraystretch}{1.12}
  \begin{tabularx}{0.55\columnwidth}{lcc}
    \toprule
    \textbf{Method} & \textbf{Sync-C} $\uparrow$ & \textbf{Conditioning Latency (s)} $\downarrow$ \\
    \midrule
    Waveform + wav2vec & \textbf{5.83} & 0.051 \\
    Single-codebook units & 2.07 & \textbf{0.003} \\
    16-codebook units & 4.95 & 0.011 \\
    \bottomrule
  \end{tabularx}
\end{table}

\paragraph{Full Inference Scaling Results.}
\label{sec:full_streaming_latency}
Table~\ref{tab:streaming_latency_full} reports results across all tested resolutions. Each setting uses the same 30 OmniCharacter requests. Additional model parallelism does not reduce the Teacher's latency under this workload. In contrast, splitting the Student
pipeline across four GPUs substantially reduces both component and end-to-end latency. The two-step Student renders faster than real time at all resolutions, while the four-step Student achieves real-time video rendering at the rectangular resolutions. The eight-GPU deployment runs two independent four-GPU replicas rather than accelerating a single request. It therefore provides similar per-request latency but doubles serving capacity. Its aggregate RTF is below 1 for both the two- and four-step Students at all tested resolutions, demonstrating real-time serving throughput under concurrent requests. Single-request E2E RTF remains slightly above 1, with the best result reaching 1.152. These measurements show that real-time end-to-end generation is within reach on higher-throughput per-replica hardware.

\begin{table*}[htbp]
  \centering
    \caption{Full inference throughput and latency results. Video FPS measures rendering throughput, and E2E RTF covers the complete omni-modal response pipeline. Rect. denotes $400\times720$/$720\times400$, Square denotes $720\times720$, and All aggregates all three shapes. The eight-GPU setting comprises two independent four-GPU replicas, so Aggregate RTF measures joint serving throughput whereas E2E RTF remains per request.}
  \label{tab:streaming_latency_full}
  \scriptsize
  \setlength{\tabcolsep}{5.45pt}
  \renewcommand{\arraystretch}{1.12}
  \begin{tabularx}{\textwidth}{cccccccccc}
    \toprule
    \textbf{GPU} & \textbf{Resolution} & \textbf{Aggregate RTF} $\downarrow$ & \textbf{E2E RTF} $\downarrow$ & \textbf{Text TPS} $\uparrow$ & \textbf{Text RTF} $\downarrow$ & \textbf{Speech units/s} $\uparrow$ & \textbf{Speech RTF} $\downarrow$ & \textbf{Video FPS} $\uparrow$ & \textbf{Video RTF} $\downarrow$ \\
    \midrule
    \rowcolor{lightgray} \multicolumn{10}{c}{\textbf{Teacher (50-step offline quality)}} \\
    1 & Rect. & \textbf{21.397} & \textbf{21.713} & \textbf{36.243} & \textbf{0.837} & \textbf{12.339} & \textbf{1.013} & 1.299 & 19.645 \\
    4 & Rect. & 26.241 & 26.917 & 7.389 & 3.770 & 2.581 & 4.844 & \textbf{1.409} & \textbf{18.086} \\
    8 & Rect. & 29.195 & 29.900 & 7.317 & 4.465 & 2.558 & 4.887 & 1.256 & 20.340 \\
    \midrule
    1 & Square & \textbf{33.949} & \textbf{34.284} & \textbf{36.216} & \textbf{0.745} & \textbf{12.371} & \textbf{1.010} & 0.779 & 32.247 \\
    4 & Square & 39.972 & 40.382 & 7.372 & 3.414 & 2.573 & 4.859 & \textbf{0.791} & \textbf{31.769} \\
    8 & Square & 41.674 & 41.778 & 7.349 & 3.916 & 2.560 & 4.883 & 0.773 & 32.643 \\
    \midrule
    1 & All & \textbf{25.861} & \textbf{25.903} & \textbf{36.234} & \textbf{0.807} & \textbf{12.350} & \textbf{1.012} & 1.125 & 23.845 \\
    4 & All & 31.148 & 31.405 & 7.384 & 3.651 & 2.578 & 4.849 & \textbf{1.203} & \textbf{22.647} \\
    8 & All & 33.608 & 33.859 & 7.327 & 4.282 & 2.559 & 4.886 & 1.095 & 24.441 \\
    \midrule
    \rowcolor{lightgray} \multicolumn{10}{c}{\textbf{Streaming Student (2-step)}} \\
    1 & Rect. & 2.758 & 2.802 & 37.388 & 0.134 & 6.752 & 1.854 & 25.692 & 0.973 \\
    4 & Rect. & 1.172 & 1.201 & 102.287 & 0.050 & 14.912 & 0.842 & 39.546 & 0.633 \\
    8 & Rect. & \textbf{0.587} & \textbf{1.152} & \textbf{133.418} & \textbf{0.038} & \textbf{15.315} & \textbf{0.819} & \textbf{40.914} & \textbf{0.611} \\
    \midrule
    1 & Square & 3.223 & 3.269 & 40.422 & 0.107 & 4.961 & 2.522 & 14.770 & 1.693 \\
    4 & Square & 1.290 & 1.317 & 84.979 & 0.057 & 14.815 & 0.847 & 26.175 & 0.955 \\
    8 & Square & \textbf{0.637} & \textbf{1.258} & \textbf{134.181} & \textbf{0.036} & \textbf{15.288} & \textbf{0.820} & \textbf{26.681} & \textbf{0.937} \\
    \midrule
    1 & All & 2.939 & 2.957 & 38.400 & 0.125 & 6.155 & 2.077 & 22.051 & 1.213 \\
    4 & All & 1.214 & 1.240 & 96.518 & 0.052 & 14.880 & 0.843 & 35.089 & 0.740 \\
    8 & All & \textbf{0.596} & \textbf{1.187} & \textbf{133.673} & \textbf{0.037} & \textbf{15.306} & \textbf{0.819} & \textbf{36.170} & \textbf{0.720} \\
    \midrule
    \rowcolor{lightgray} \multicolumn{10}{c}{\textbf{Streaming Student (4-step)}} \\
    1 & Rect. & 2.936 & 2.977 & 42.116 & 0.119 & 5.936 & 2.109 & 19.105 & 1.309 \\
    4 & Rect. & 1.265 & \textbf{1.293} & 109.626 & 0.046 & 14.917 & 0.841 & \textbf{26.512} & \textbf{0.943} \\
    8 & Rect. & \textbf{0.655} & 1.299 & \textbf{134.343} & \textbf{0.037} & \textbf{15.298} & \textbf{0.820} & 25.863 & 0.967 \\
    \midrule
    1 & Square & 3.789 & 3.834 & 42.101 & 0.103 & 4.050 & 3.089 & 10.775 & 2.321 \\
    4 & Square & 1.781 & 1.800 & 93.474 & 0.051 & 14.905 & 0.841 & 16.891 & 1.480 \\
    8 & Square & \textbf{0.891} & \textbf{1.769} & \textbf{134.325} & \textbf{0.036} & \textbf{15.275} & \textbf{0.821} & \textbf{16.958} & \textbf{1.475} \\
    \midrule
    1 & All & 3.268 & 3.263 & 42.111 & 0.113 & 5.307 & 2.436 & 16.328 & 1.646 \\
    4 & All & 1.447 & 1.462 & 104.242 & 0.048 & 14.913 & 0.841 & \textbf{23.305} & \textbf{1.122} \\
    8 & All & \textbf{0.735} & \textbf{1.455} & \textbf{134.337} & \textbf{0.037} & \textbf{15.291} & \textbf{0.820} & 22.895 & 1.136 \\
    \midrule
    \rowcolor{lightgray} \multicolumn{10}{c}{\textbf{Streaming Student (8-step)}} \\
    1 & Rect. & 3.615 & 3.656 & 42.216 & 0.118 & 4.557 & 2.747 & 12.502 & 2.000 \\
    4 & Rect. & 1.903 & \textbf{1.932} & 111.752 & 0.046 & 14.935 & 0.840 & \textbf{15.622} & \textbf{1.601} \\
    8 & Rect. & \textbf{0.986} & 1.944 & \textbf{133.526} & \textbf{0.038} & \textbf{15.517} & \textbf{0.808} & 15.332 & 1.631 \\
    \midrule
    1 & Square & 5.026 & 5.074 & 42.704 & 0.101 & 2.925 & 4.276 & 6.978 & 3.584 \\
    4 & Square & 2.900 & 2.912 & 98.991 & 0.049 & 14.895 & 0.842 & 9.628 & 2.598 \\
    8 & Square & \textbf{1.439} & \textbf{2.844} & \textbf{134.561} & \textbf{0.036} & \textbf{15.494} & \textbf{0.809} & \textbf{9.798} & \textbf{2.552} \\
    \midrule
    1 & All & 4.164 & 4.129 & 42.379 & 0.113 & 4.013 & 3.257 & 10.661 & 2.528 \\
    4 & All & 2.256 & 2.258 & 107.499 & 0.047 & 14.922 & 0.841 & \textbf{13.624} & \textbf{1.933} \\
    8 & All & \textbf{1.132} & \textbf{2.244} & \textbf{133.871} & \textbf{0.037} & \textbf{15.509} & \textbf{0.809} & 13.487 & 1.938 \\
    \bottomrule
  \end{tabularx}
\end{table*}

\paragraph{Human Preference Evaluation against Recent Methods.}
We conduct a blind human preference study on 200 CommonEval requests,
comparing Ex-Omni-2D with OmniAvatar and UniAVGen. We recruit five professional annotators with extensive experience in video-caption annotation and speech/ASR annotation. For each request, annotators rank the anonymized outputs in terms of lip synchronization and facial-motion naturalness, with ties allowed. Table~\ref{tab:human_method_comparison} reports the mean rank and first-tier rate. OmniAvatar achieves the strongest preference on both criteria. Ex-Omni-2D and UniAVGen show complementary strengths: UniAVGen performs better on lip synchronization, while Ex-Omni-2D obtains a slightly better mean rank for facial naturalness. The two methods therefore show broadly comparable overall performance, but both remain behind OmniAvatar. This gap also reflects the advantage of OmniAvatar as a model specialized for audio-driven avatar rendering, whereas Ex-Omni-2D addresses unified dialogue, speech, and video response generation.

\begin{table}[htbp]
\centering
\caption{Blind human comparison with recent methods. Each cell reports mean
rank / first-tier rate (\%). Lower mean rank is better; first-tier rates
include ties.}
\label{tab:human_method_comparison}
\scriptsize
\setlength{\tabcolsep}{6pt}
\begin{tabular}{lcc}
\toprule
\textbf{Method} & \textbf{Lip Sync} & \textbf{Facial Naturalness} \\
\midrule
OmniAvatar-1.3B & \textbf{1.290 / 75.6} & \textbf{1.396 / 64.9} \\
UniAVGen         & 1.683 / 48.4 & 1.932 / 29.9 \\
Ex-Omni-2D       & 1.755 / 42.6 & 1.866 / 32.5 \\
\bottomrule
\end{tabular}
\end{table}

\paragraph{Human Preference across Guidance Scales.}
We further conduct a blind comparison of five Teacher guidance settings on the same 200 CommonEval requests, using the same five professional annotators. Annotators rank the outputs by lip synchronization, facial-motion naturalness, and gesture quality, with ties allowed. As shown in Table~\ref{tab:human_cfg_comparison}, CFG~4 / Audio CFG~4 achieves the best human preference for lip synchronization and facial naturalness while retaining nearly the same gesture rank as CFG~8 / Audio CFG~8. Although stronger guidance can improve automatic Sync-C in some settings, annotators often perceive the resulting motion as exaggerated and less natural. This reveals a gap between automatic synchronization metrics and human preference: Sync-C captures audiovisual correspondence but does not fully reflect motion naturalness. Increasing Audio CFG alone also fails to improve perceived lip synchronization. 

\begin{table}[htbp]
\centering
\caption{Human evaluation of Teacher guidance scales. Each cell reports mean
rank / first-tier rate (\%). Lower mean rank is better; first-tier rates
include ties.}
\label{tab:human_cfg_comparison}
\scriptsize
\setlength{\tabcolsep}{4pt}
\begin{tabular}{cccc}
\toprule
\textbf{CFG / Audio CFG} & \textbf{Lip Sync}
& \textbf{Facial Naturalness} & \textbf{Gesture} \\
\midrule
1 / 1     & 3.226 / 14.7 & 2.565 / 27.5 & 2.013 / 47.7 \\
2 / 2     & 2.564 / 23.2 & 2.187 / 35.8 & 1.763 / 56.5 \\
3.5 / 8.5 & 2.275 / 27.3 & 2.510 / 20.3 & 1.700 / 57.0 \\
4 / 4     & \textbf{1.780 / 50.0} & \textbf{1.863 / 45.7}
          & 1.580 / 63.1 \\
8 / 8     & 2.530 / 21.4 & 2.985 / 13.8
          & \textbf{1.571 / 63.9} \\
\bottomrule
\end{tabular}
\end{table}

\end{document}